\documentclass[final,5p,times,twocolumn]{elsarticle}

\usepackage{verbatim} %multiline comment 

\usepackage{graphicx}
\usepackage{subcaption}
\usepackage{amssymb}
\usepackage{amsmath}
\DeclareMathOperator*{\argmin}{arg\min \ }
\usepackage{lineno}

\usepackage{multirow}
\usepackage[table]{xcolor}
\usepackage{pgfplots}
\pgfplotsset{width=6.5cm,compat=1.14}
\usepackage{hyperref}
\biboptions{round,colon,authoryear}

\journal{Computerized Medical Image and Graphics (CMIG)}
\newcommand{\rev}[1]{{#1}}

\newcommand{\bib}[1]{(\cite{#1})} 

\begin{document}

\begin{frontmatter}


\title{From Patch to Image Segmentation using Fully Convolutional Networks - 
        Application to Retinal Images}

% \author{Taibou Birgui Sekou \corref{cor1}\corref{cor2}\corref{cor3}}
%\ead{taibou.birgui\_sekou@insa-cvl.fr}
%\cortext[cor1]{Institut National des Sciences Appliqu\'{e}es Centre Val de Loire, Blois France}
%\cortext[cor2]{Universit\'{e} de Tours, Tours, France}
%\cortext[cor3]{ LIFAT EA 6300, Tours, France}

 \author[label1,label2,label3,label4]{Taibou Birgui Sekou}
 \author[label1,label2,label3]{Moncef Hidane}
 \author[label1,label2,label3]{Julien Olivier}
 \author[label2,label3]{Hubert Cardot}
 \address[label1]{Institut National des Sciences Appliqu\'{e}es Centre Val de Loire, Blois France}
\address[label2]{Universit\'{e} de Tours, Tours, France}
\address[label3]{LIFAT EA 6300, Tours, France}
\address[label4]{Corresponding author: taibou.birgui\_sekou@insa-cvl.fr}

%\author{Taibou Birgui Sekou}

%\address{Tours, France}

\begin{abstract}
Deep learning based models, generally, require a large number of samples for appropriate training, a 
requirement that is difficult to satisfy in the medical field.
This issue can usually be avoided with a proper initialization of the weights.
On the task of medical image segmentation in general, two techniques are oftentimes employed to
tackle the training of
a deep network $f_T$. The first one consists in reusing some weights of a network $f_S$ pre-trained
on a large scale database ($e.g.$ ImageNet). This procedure, also known as \textit{transfer learning}, happens to
reduce the flexibility when it comes to new network design since $f_T$ is constrained to match some 
parts of $f_S$. The second commonly used technique consists in working on image patches to benefit from
the large number of available patches. This paper brings together these two techniques and propose to train
\emph{arbitrarily designed networks} that segment an  image in one forward pass, with a focus on relatively small databases.
% in two stages: patch pre-training and  full sized image fine-tuning.   
An experimental work have been carried out on the tasks of retinal blood vessel segmentation and
the optic disc one, using
four publicly available databases. Furthermore, three types of network are considered, going 
from a very light weighted network to a densely connected one. The final results show the efficiency of the proposed
framework along with state of the art results on all the databases. 
The source code of the experiments is publicly available at \url{https://github.com/Taib/patch2image}.
%%%%
\end{abstract}


\end{frontmatter}


\section{Introduction}
\label{S:1}

\begin{figure}[t] 
    \centering  
  \centerline{\includegraphics[width=8cm]{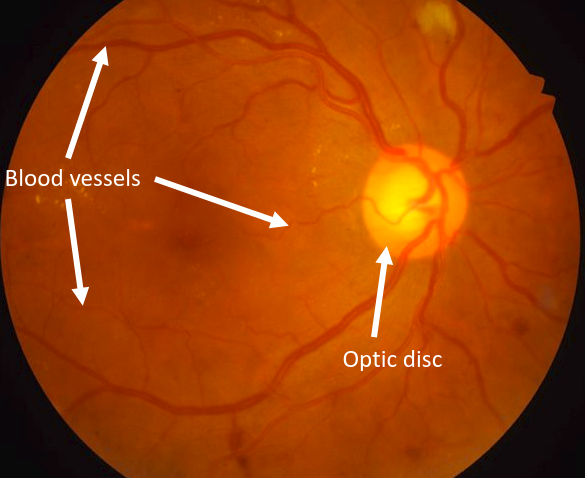}}
    \caption{Illustration of a retinal image and some structures therein. (For interpretation of the
    references to color in this figure legend, the reader is referred to the web version
    of this article.)}
    \label{fig: wiw_retina} 
\end{figure}

The task of image segmentation consists in partitioning an input picture into non-overlapping regions. On medical images,
it aims at highlighting some regions of interest (ROIs) for a more in-depth study/analysis. 
The result of a segmentation can then be used to extract various
morphological parameters of the ROIs that bring more insights into the diagnosis or control of diseases. 
On a retinal image, the ROIs include the blood vessels and the optic disc 
(the optic nerve head), as depicted in Figure \ref{fig: wiw_retina}.
Retinal blood vessel segmentation (RBVS) is helpful in the diagnosis, treatment and 
monitoring of diseases such as the diabetic retinopathy, the hypertension and the arteriosclerosis 
\bib{Bowling_2016, Abramoff_2010}. Optic disc segmentation (ODS) can bring relevant insights
on the glaucoma disease \bib{Bowling_2016}. Nowadays, these diseases are the leading factors of blindness
\bib{Yau_2012_Global, Tham_2014_Global}.\\
The manual annotation of the blood vessels (or the optic disc) is an arduous and time consuming work that requires barely-accessible expertise. Moreover, not only do expert segmentations are subject to inter- and 
intra-operator variability, they also become impractical for large-scale and real-time uses. Hence the large amount of work in designing automatic  methods \bib{Fraz_2012_blood, Srinidhi_2017_Recent}. In the following, retinal image segmentation (RIS) refers to 
either blood vessel or optic disc segmentation.

More generally, medical image segmentation  has undergone tremendous advances which are mainly due to deep 
convolutional neural networks (DCNNs). A DCNN is a mathematical model that applies a
series of feature extraction operators ($a.k.a.$ the convolution filters) on the input to obtain a high level 
abstract representation. The latter can then be used as input to a classifier/regressor
\bib{LeCun_1998_Gradient}. The model can be trained in a supervised manner using a training set. 
When employing DCNNs, image segmentation can be cast as a classification task where the goal is to label each pixel
of the input image. 

On the task of RIS, the DCNN-based propositions can be grouped by their type of input/output: patch-to-label, 
patch-to-patch, and image-to-image. The last two groups are also known as end-to-end approaches 
since their inputs and outputs are of the same spatial size. They are, almost always, based on 
fully convolutional neural networks (FCNNs) \bib{Long_2015_CVPR} which conserve the spatial 
information of their inputs throughout the computation. 

The patch-to-label group trains a network that takes an image patch (small window inside the image) as input and 
produces the  label of its center pixel. A natural variation consists in assigning labels to a \emph{subset} of the input patch. 
Then, given a test image, one classifies each pixel using a surrounding patch. Thus, the segmentation
can be cast as an independent patch classification task. 
Besides being time consuming, these approaches usually fail when the objects are on the frontiers of 
two classes. This is mainly due to the loss of contextual/structural information in the final layers.

These drawbacks can be alleviated by using a patch-to-patch network. It consists in using a 
network that can output a segmentation map for each given patch. In other words, the network labels each pixel of the 
patch. As a consequence the segmentation task is performed in three stages: a patch extraction from the input image
followed by a pass through the network to obtain the segmentation maps, finally these maps are aggregated to output 
the final segmented image. 
These approaches reduce the previous frontiers' objects issue though the patch aggregation stage may still be time 
consuming especially for real-time applications.

Note that the patch-based methods require a certain level of discriminative information in the patches, for them to be 
efficient. When this requirement is hard to fulfill ($e.g.$ the ROIs occupy the quasi-totality of the image), it might be
beneficial to work 
directly on the full-sized images. The image-to-image networks take the entire image as input and produce the complete 
segmentation in one forward pass through the network. They have the advantage of being well adapted for real-time applications and can be seen
as a generalization of the patch-to-patch approach with a large patch size.

Regarding the training of DCNNs, an important effort needs to be dedicated to building the corresponding training sets.
In general, a well built training set satisfies the following: $(i)$ it contains different examples of each sought 
invariance,  and $(ii)$ it consists of a large number of samples.
Unfortunately, these requirements are often hard to meet in the medical field: medical images are hard to obtain 
and very difficult to annotate. Nevertheless, DCNN-based models have been successfully applied in the medical field
\bib{Litjens_2017_Survey, Zhou_2017_Deep}, in particular on the task of RIS \bib{Fraz_2012_blood, Srinidhi_2017_Recent}.

The first trick used in training DCNNs for RIS is to work at the patch-level ($i.e.$ using either a patch-to-label approach 
or a patch-to-patch one). 
Allowing for patch overlap when extracting training patches leads to large-scale training sets, thus facilitating the 
training of arbitrarily designed
networks \bib{Liskowski_2016_Segmenting, Li_2016_Cross}.%, on the grounds that patches contain enough discriminative information. 

When working at the image-level ($i.e.$ using an image-to-image approach) two tricks are usually combined to train
DCNNs on small sized databases. 
The first one, known as \emph{data augmentation}, consists in adding to the initial database scaled and rotated versions of the training samples. It helps 
boosting the training set and to some extent avoiding over-fitting.
Yet, the training set may still be very small compared to the number of parameters in the model.
The second technique, known as 
\emph{transfer learning}, consists in reusing the parameters of a network $f_S$ trained on a source 
database $\mathcal{D}_S$
to improve the performance of a network $f_T$ to be learned on a target database $\mathcal{D}_T$, as depicted in 
Figure \ref{fig: tlj}. 
Consequently, the architecture of 
the network $f_T$ is somehow predefined by the one of $f_S$ since,  
for $f_T$ to reuse some weights 
of $f_S$, the architecture of the two networks must match at least at the shared weights locations. 
Transfer learning aims at finding a good starting point for the weights and helps training very deep networks
on relatively small datasets \bib{Tajbakhsh_2016_Convolutional, Yosinski_2014_How}.

Generally, on the task of RIS, the proposed networks so far are either trained at the patch-level or at the image-level. 
Those that are trained at the patch-level can benefit from a freely designed architecture, whereas those that are trained at
the image-level should conform to previously defined architectures (usually the VGG of \bib{Szegedy_2014_Going}).
This paper presents a training framework that bridges the gap between the two levels and provides a way to have
freely designed networks working efficiently both at the patch- and image-levels even on small sized databases.
The liberty in the network design opens up the door for more research on deep architectures for RIS, with the hope of reaching findings as remarkable as those obtained for natural images such as AlexNet \bib{Krizhevsky_2012_ImageNet},
VGG \bib{Szegedy_2014_Going}, ResNet \bib{He_2015_Deep} and DenseNet \bib{Huang_2017_Densely}. It also provides a way to build budget-aware architectures \bib{Saxena_2016_Convolutional, Veniat_2018_CVPR} that can suit different
clinical constraints ($e.g.$ lightweight networks for embedded systems).

\textit{\rev{How to obtain state-of-the-art performances at the image-level using a freely designed network on the task of RIS? }} 
If the size of the database at hand is consequent then the challenge is in building an appropriate network 
and pre-processing the data. This case has been well studied in the natural image literature.
However, for small sized database, generally encountered in the medical field (especially on RIS tasks), \rev{existing networks for an image-level segmentation are generally not freely designed}.
The focus on freely designed image-to-image models leads to fast and easy to use domain specific architectures.
Thus, in this paper the small sized database case is elaborated. The followings are presented:

\begin{itemize}
\item The proposed framework consists of four phases. The first two take advantage 
of the large-scale training samples available at the patch-level to train the networks from scratch
($i.e.$ from random initialization).
The last two phases consist of a transfer learning from the patch networks
weights to the image ones and a fine-tuning to adapt to weights and improve the segmentation.

\rev{It is worth noting that when employing a FCNN as a patch-to-patch segmentation
model, it can naturally take as input the entire image since FCNNs are invariant to the input 
spatial size. Therefore, the proposed framework can be seen as an analysis of how transferable
are the weights of a patch-to-patch FCNN when applied to the entire images and the importance of 
an additional fine-tuning at the image-level.}
%The trained networks, which work quite well at the patch-level, 
%are then fine-tuned at the image-level. 

%An experimental study is carried out on how transferable are the weights from the patch-level to 
%the image-one and how to perform the transfer. 
\item To assess the results of the proposed framework, 3 types of network are considered. 
The networks vary in terms of architectural complexity and number of weights: $(1)$
a lightweight network (Light) of $8,889$ weights, $(2)$ a reduced UNet-like \bib{UNET} network (Mini-Unet)
of $316,657$ weights, and $(3)$ a deep densely connected one (Dense) of $1,032,588$ parameters
inspired from \bib{Huang_2017_Densely}. 
\item Concerning the databases, the experiments have been carried out on four publicly available databases with a
relatively small number of training samples. On the task of RBVS, three databases are used for evaluation:
DRIVE, STARE, and CHASE\_DB1. Whereas the IDRID is used for the ODS task.
\item The quantitative results show state-of-the-art results of the final image-to-image networks.
Furthermore, they prove that all the patch-based networks can be extended to work directly at the 
image-level with a  performance improvement. All the phases of the proposed framework 
are analysed to point out their importance. On the other hand, using the framework,
we show that on some databases the Light network can perform as good as other complex 
state-of-the-art networks while staying by far more simple and segmenting an entire image in one forward pass. 
\end{itemize}

The rest of the paper is organized as follows. Sec. \ref{S:rw} gives an overview of the related work. After
a brief introduction to DCNNs (Sec. \ref{S:cnn}) and transfer learning (Sec. \ref{S:tl}),
Sec. \ref{S:fw} presents the proposed framework. 
Sec. \ref{S:ex} presents the experimental work including the used DCNNs, the databases and their 
associated pre-processing along with the results and discussions.
We summarize the proposed work in Sec. \ref{S:sum} and provide some perspectives.

\section{Related Work}
\label{S:rw} 
In the following, some recent deep learning based propositions on RIS are listed and grouped by their
training strategy: patch- and image-based.  For a more detailed review see \bib{Fraz_2012_blood, Srinidhi_2017_Recent, Moccia_2018_Blood}.

\subsection{Patch-based training} 
On the task of RBVS, \citet{Liskowski_2016_Segmenting} proposed a freely designed  patch-to-label DCNN which is further
extended to output a small patch centered on the input. The network is trained from scratch using $27\times 27$ RGB 
patches. In \bib{Hajabdollahi_2018}, an additional complexity constraint is imposed on the patch-to-patch network.
The goal was to build networks that work efficiently on real-time embedded systems, for example
a binocular indirect ophthalmoscope. \citet{Oliveira_1018_Fully} proposed a patch-to-patch model and
added a stationary wavelet transform pre-processing step to improve the network's performance. On the
other hand, \bib{Jiang_2018_Retinal} chose to use plain RGB patches but reused the 
AlexNet network  \bib{Krizhevsky_2012_ImageNet} which is pre-trained on the ImageNet dataset.
Other propositions include \bib{Dasgupta_2017_Fully, Yang_2017_Patch, Li_2016_Cross}, where
 freely designed patch-to-patch networks are once again trained from scratch.

On the task of optic disc segmentation, Fu et al. \bib{Fu_2018_Joint} presented a multi-level FCNN with a 
patch-to-patch scheme which aggregates segmentations from different scales. The network is applied 
on patches centered on the optic  disc. Therefore, an optic disc detection is required beforehand. 
\citet{Tan_2017_Optic} proposed a patch-to-label network which classifies the central pixel of 
an input patch into four classes: blood vessel, optic disc, fovea, and background.
Because of the lack of training data, most of the RIS models are patch-based. However, some propositions
have been made using an image-to-image scheme based on weights transfer from the VGG network.

\subsection{Image-based training} 
In a more general setting of medical image segmentation, \citet{Tajbakhsh_2016_Convolutional} showed
experimentally that training with transfer learning performs at least as good as training from scratch.
On the task of RBVS, \citet{Fu_2016} used the image-to-image HED network \bib{HED} and added 
a conditional random field post-processing step. They performed a fine-tuning from the HED which 
is based on the VGG of \citet{Simonyan_2014_Very}. \citet{Mo_2017_Multi} proposed to reuse the VGG and
added supervision at various internal layers to prevent the gradient from vanishing in the shallower layers. 
A similar technique has been used by \citet{Maninis_2016} with an additional path to segment both the 
blood vessels and the optic disc. 

\subsection{From patch- to image-based training} 

A proof of concept is presented in our previous work \bib{Birgui_2018_Retinal}. Promising results  
have been obtained when using the patch network's weights as an initialization point of the image network.
However the patch network's metrics tend to be better than the image ones. This remark may question the 
importance of the transfer. Moreover, only one database and a single network have been considered. 
Hence, one may question the generalization to other databases and networks. 

This paper redefines the framework in a more detailed manner and applies it to three types of networks.
The networks are trained with a segmentation-aware loss function using decay on the learning rate
to avoid the over-fitting observed in the previous work when fine-tuning at the image-level.
Furthermore, the experiments have been carried out on four publicly available databases. The final
results prove that the results of the patch network can be improved using the framework.

\begin{figure*}[t] 
    \centering 
  \centerline{\includegraphics[width=13cm]{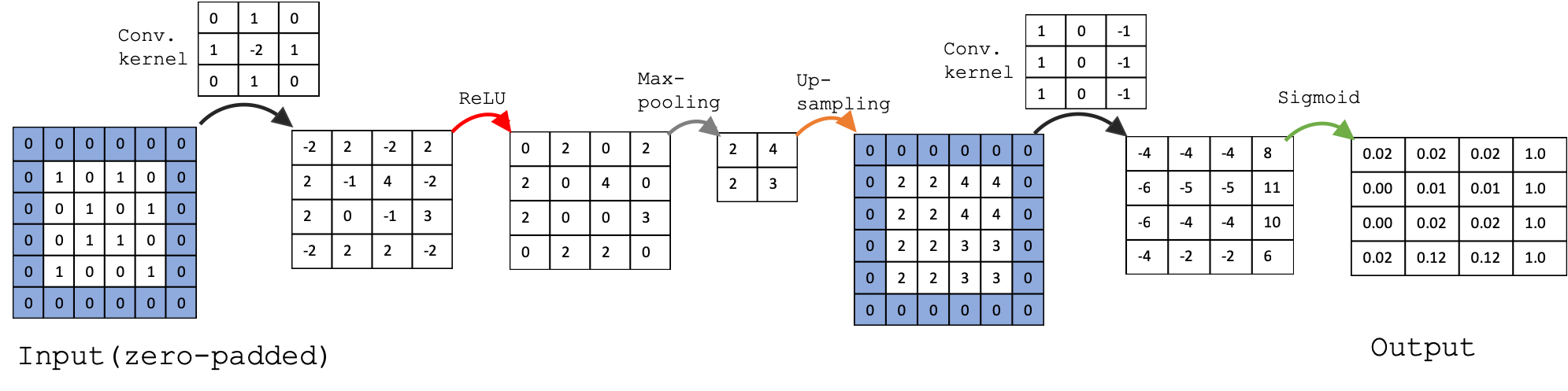}}
    \caption{A toy example FCNN. The input and output are of size $4\times 4$. The highlighted blue cells 
    represents the zero-padding applied before each convolution to maintain the input size. The convolution
    kernels are given on top of the black arrows. The red arrow corresponds to the point-wise rectified
    linear unit function: $x \mapsto \max(x, 0)$. The gray arrow applies a max-pooling in $2\times 2$ windows
    with a stride of $2$. The orange arrow is an up-sampling layer that duplicate the values in a $2\times 2$ window.
    The green arrow consists in applying the sigmoid function, point-wise. (For interpretation of the
    references to color in this figure legend, the reader is referred to the web version
    of this article.)}
    \label{fig: fcn_ex} 
\end{figure*}
  
\section{Convolutional Neural Networks}
\label{S:cnn}
With a first success in the late 80's \bib{LeCun_1989_Backpropagation}, DCNNs are nowadays the method of choice 
in diverse areas, particularly in computer vision including image classification 
\bib{He_2015_Deep, Huang_2017_Densely, Simonyan_2014_Very}, image segmentation \bib{Long_2015_CVPR}, image generation 
\bib{Goodfellow_2014_Generative}, to name a few. 
As stated in the introduction, DCNNs for classification consist in extracting highly abstract and hierarchical
representations from the 
data that are fed into a classifier/regressor. They consist of different types of layers that are stacked on top of each
other to form the final model. Commonly used layers include: the convolution layer, non-linearity (ReLU, Sigmoid, ELU), 
normalization (batch-normalization), pooling (max-pooling, average-pooling), dropout, fully connected, etc.
The convolution and fully connected layers aim at learning high-level patterns from the database that
can improve the representation. 
On the other hand, pooling, non linearity and normalization layers aim at adding more invariance 
into the representation and ameliorating the generalization power. A clear view of these models can be found in 
\bib{Goodfellow_2016_Deep, LeCun_2015_Deep, Zeiler_2014_Visualizing}.

On the task of image segmentation, state-of-the-art DCNN-based models are usually fully convolutional 
neural networks (FCNNs). FCNNs are a variation of DCNNs that do not include fully connected layers \bib{Long_2015_CVPR}. Hence, they allow the 
spatial/structural 
information of the input to be preserved across the network. As a consequence, one can build end-to-end like networks where 
the output and the input are of the same shape. Therefore one can directly train a network to segment an entire image 
in one forward pass. Additional layers that appear in FCNNs, used 
to recover the size of the input, are the transposed convolution ($a.k.a$ deconvolution) layer and the up-sampling 
($a.k.a$ duplication) one.
Another advantage of the FCNNs, inherited from the convolutions, is the input size invariance. That is, a same FCNN can
be applied to inputs of various size as long as they all share the same number of spatial dimensions and channels. For
example, the FCNN of Figure \ref{fig: fcn_ex} can, as well, take inputs of shape $400\times 400$ to output a $400\times 400$ image.

Hereafter, a convolutional network $h$ is characterized by its associated set of parameters $\theta$. Typically, if $h$ is FCNN
then $\theta$ will be of the following form:
\begin{equation}
\label{eq: cnn}
 \theta = \{\{W_{l,k}, b_{l,k}\}_{k=1}^{K_l}, \sigma_l\}_{l=1}^L,
\end{equation}
where $L$ is the number of layers, $K_l$ is the number of kernels in the $l$-th layer, $\sigma_l$ is the $l$-th activation function, 
$W_{l,k}$ is the $k$-th kernel (the convolution kernel) of the layer $k$, and 
$b_{l,k}$ is the associated bias.

\begin{figure}[t] 
    \centering  
  \centerline{\includegraphics[width=8cm]{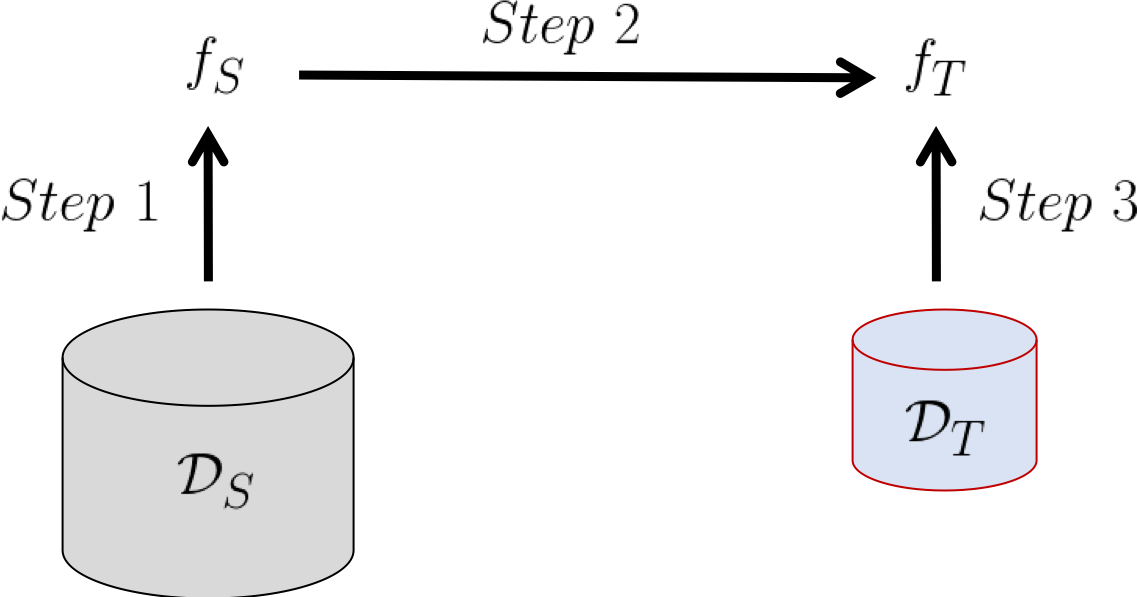}}
    \caption{Classical transfer learning. 
        Step 1:  Train the network $f_S$ on a large scale database ($e.g.$ ImageNet). 
        Step 2:  Build a task specific network $f_T$ that reuse some weights of $f_S$. 
        Step 3:  Fine-tune the network $f_T$ on $\mathcal{D}_T$.  }
    \label{fig: tlj} 
\end{figure}
\section{Transfer Learning}
\label{S:tl}
In the case of DCNNs, transfer learning consists in reusing the ``knowledge'' learned by a network $N_S$ trained on a source
database $\mathcal{D}_S$ to ameliorate the performance of a network $N_T$ on a target database $\mathcal{D}_T$, with $\mathcal{D}_S \neq \mathcal{D}_T$ 
($see$ Figure \ref{fig: tlj}). Here a database $\mathcal{D}_i$ consists of two elements: the input data $\mathbf{X}_i$ 
and their targets $\mathbf{Y}_i$. Thus $\mathcal{D}_S \neq \mathcal{D}_T$ means $(1)$ $\mathbf{X}_S \neq \mathbf{X}_T$ 
and/or $(2)$ 
$\mathbf{Y}_S \neq \mathbf{Y}_T$. In other words, either $(1)$ the inputs are different ($e.g.$ natural images $v.s.$ 
medical images, patches $v.s.$ images) or $(2)$ the targets are different ($e.g.$ retinal blood vessels $v.s.$ 
optic discs).

The knowledge in a DCNN lies in its weights ($i.e.$ the convolution filters) learned throughout
the training. DCNNs based transfer learning can be seen as a \emph{feature-representation-transfer} and a
\emph{hyper-parameter-transfer}.
It is a hyper-parameter-transfer since transferring some knowledge from
$N_s$ to $N_t$ implies the reuse of part of the architecture of $f_S$ in $f_T$. 
The architecture per se of a DCNN is a hyper-parameter.
Concerning the feature-representation-transfer, the transferred weights will write inputs from $\mathcal{D}_S$ and $\mathcal{D}_T$
in the same feature space. Therefore, the weights' transfer constrains the two networks to share the
same feature representation.

There are mainly three major concerns when transferring knowledge of DCNNs: what to transfer, when to do it, and 
how to perform it \bib{Pan_2010_Survey}. Answering these questions helps avoiding a \emph{negative transfer},
that is, degrading the performance on the target database.\\
What to transfer is equivalent to knowing which weights to transfer. In other words, which part of the architecture of $f_S$
will be reused in $f_T$. While the weights in the first layers of a DCNN capture abstract patterns of the data, the ones in
the deeper layers (especially the last layer) are more task specific.
The work of \citet{Yosinski_2014_How} provides a more detailed experimental work on how much abstracted are 
the learned weights.\\
The when to transfer question asks which couple ($\mathcal{D}_S$, $f_S$) is well suited to transfer from. 
For example, the current state-of-the-art results on transfer 
learning advise to have at least two databases of the same types ($i.e.$ image to image, audio to audio).\\
The last question (how to transfer) is resolved mainly with a copy-paste of the weights from one network to another.
Then, one can either keep the transferred weights \emph{frozen} or  \emph{fine-tune} them. If frozen, the
weights are not updated, while,  the fine-tuning case continues to update the weights using the target database.
In the latter case, transfer learning is equivalent to weight initialization. 

Most DCNNs based RIS systems that perform transfer learning reuse
the convolution layers of the VGG network and add some additional task specific layers. As a consequence, the final network's
architecture is no longer freely designed. Often, RIS may not need an extremely sophisticated network to achieve
good results. In the following, a proposed framework to train a freely designed 
network on RIS tasks is presented.

\begin{figure*}[t]  
    \centering
  \centerline{\includegraphics[width=14cm]{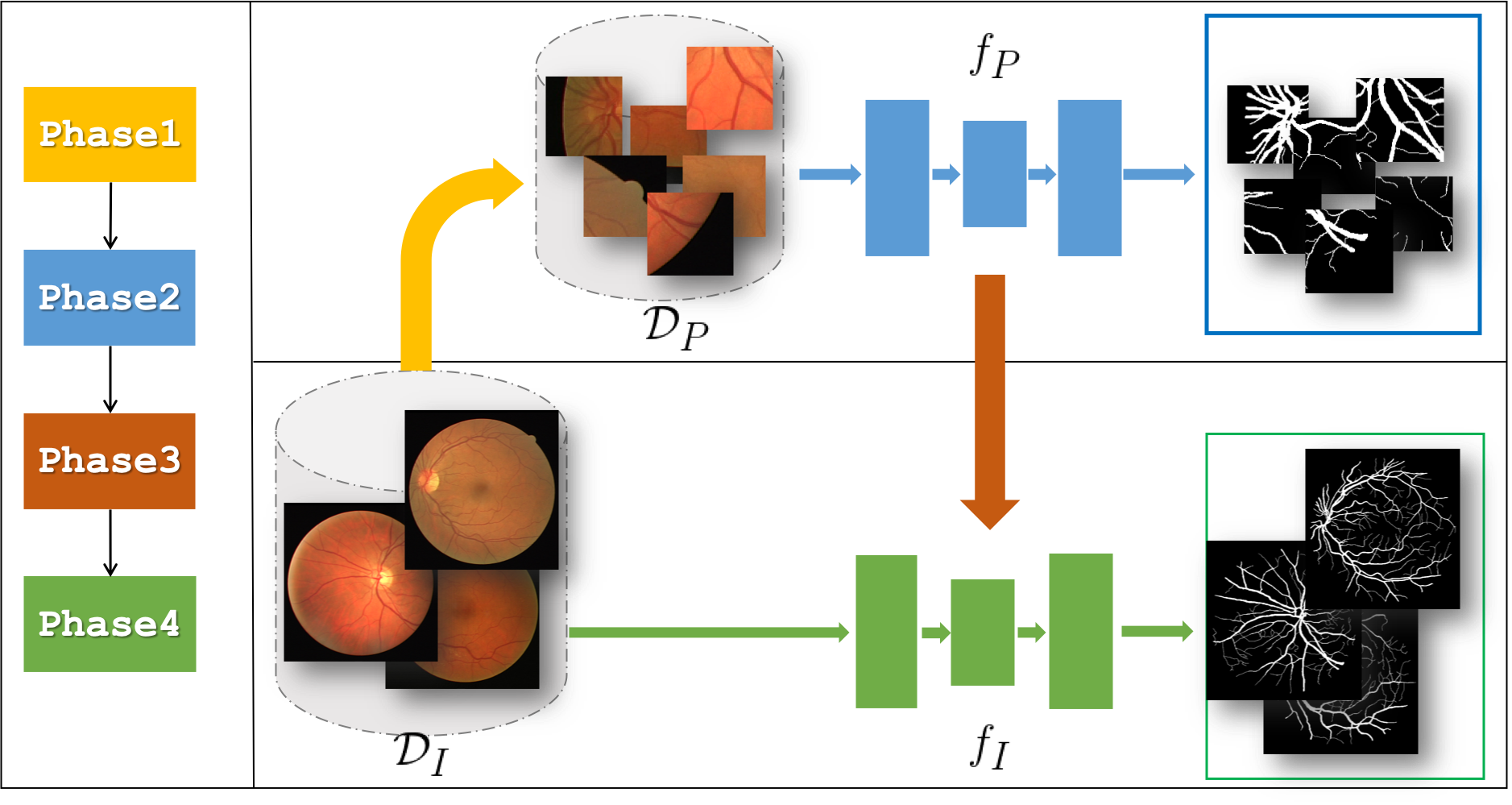}}
    \caption{Flowchart of the proposed framework. $\mathcal{D}_I$ is the database of the original full-sized images. Patches are extracted
    from the images of $\mathcal{D}_I$ to build $\mathcal{D}_P$. $f_P$ is the network trained on $\mathcal{D}_P$, while $f_I$
    is the one trained on $\mathcal{D}_I$. (For interpretation of the
    references to color in this figure legend, the reader is referred to the web version
    of this article.)}
    \label{fig: fwj}  
\end{figure*}

\section{The Framework}
\label{S:fw}
Given a database $\mathcal{D}_I$ of retinal images (and their corresponding annotations) 
our principal task is to freely design 
and train a FCNN efficiently. As aforementioned, for the considered RIS databases and the medical field in general,
$\mathcal{D}_I$ usually 
is of relatively small size which makes difficult the training of complex FCNNs architecture. 
This issue is commonly addressed by working at the patch-level \bib{Zhou_2017_CVPR}. 
Therefore, the proposed framework makes use of the patch-level information to build large-scale patch 
database. Figure \ref{fig: fwj} depicts the four main components of the framework which are detailed in the 
following. 
Before detailing the components, we start by given some theoretical motivation of the framework.

\subsection{Theoretical Motivation}

For the sake of notation simplicity, we consider a binary image segmentation task with square-shaped images with one channel.
Namely, let $\mathcal{I} \subset \mathcal{M}_d(\mathbb{R})$ be the $d^2$-dimensional image space and
$\mathcal{S} \subset \mathcal{M}_d(\{0,1\}) $ be the associated segmentation space such that  there exists an oracle $f: \mathcal{I} \to \mathcal{S}$
that assigns to each image $X\in \mathcal{I}$ its unique segmentation $Y=f(X)$. Suppose the images are sampled from a distribution $\mathbb{P}_{x}$, whose support is precisely the set $\mathcal{I}$.
The objective is to approximate $f$ with a parametrized function $\tilde{f}$ that minimizes the expected loss 
$\mathbb{E}_{X \sim \mathbb{P}_{x}}\Big[ r(\tilde{f}(X), f(X))\Big]$,
where $r$ is a certain elementary loss function. In practice, the expectation is approximated by the empirical mean 
over a dataset $\mathcal{D} = \{X_i, Y_i=f(X_i)\}_{i=1}^n$ of $n$ samples with $X_i \in \mathcal{I}$
and $Y_i \in \mathcal{S}$. In other words, $\tilde{f}$ is given by
\begin{equation}
\label{eq: approx}
\tilde{f} = \argmin_{h \in \mathcal{H}}(1/n) \sum_{i=1}^n r(h(X_i), f(X_i)),
\end{equation}
where $\mathcal{H}$ is the class of parameterized functions in use. In our case, $\mathcal{H}$ is the set of functions that correspond to a forward pass through our neural network. On the task of RIS, one might want to start the segmentation locally (in smaller neighborhoods) then
fine-tune it by looking at the entire image. Using this idea, the optimization problem of Eq. \ref{eq: approx} involving $d^2$-dimensional $\mathcal{I}$ images now involves $p^2$-dimensional patches, where $p$ is the size of neighborhoods. This introduces the use of patch-based approaches that are discussed in the following.

An image $X \in \mathcal{I}$ can be written as a combination of $p^2$-dimensional patches by constructing a set 
of functions $\{\nu_k, \varepsilon_k \}_{k=1}^P$, with $P$ the number of patches, such that: 
\begin{equation}
\label{eq: patch}
X = \sum_{k=1}^P \nu_k(\varepsilon_k(X)) ,
\end{equation}
where, for all $k$, $\varepsilon_k: \mathcal{I} \to \mathcal{M}_p(\mathbb{R})$, and $\nu_k: \mathcal{M}_p(\mathbb{R}) \to \mathcal{I}$. 
The function $\varepsilon_k$ extracts the $k$-th patch while $\nu_k$ plugs the latter in the image and handles the possible overlap. In Eq. \ref{eq: patch}, we are assuming that each pixel of $X$ can be 
reconstructed from the patches it belongs to.
For instance, if we consider non-overlapping patches of $\mathcal{M}_2(\mathbb{R})$ and $d=6$ then 
\[
 X = \sum_{r=0}^{2} \sum_{s=0}^{2} A_r^T A_r X A_s^T A_s,
\]
where $A_k = A_{k-1}J_{0,6}^2 \in \mathcal{M}_{2,6}(\{0, 1\})$ for all $k\geq 1$, $J_{0,6}$ is the $6\times 6$ Jordan  matrix with $0$ as eigenvalue
and \[A_0 = \begin{bmatrix}
    1 & 0 & 0 &0& 0 & 0 \\
    0  & 1 & 0 & 0& 0 & 0 \\
\end{bmatrix} \in \mathcal{M}_{2,6}(\{0, 1\}). \]
So for a fixed $r$ and $s$, the matrix $A_r X A_s^T$ is an element of $\mathcal{M}_2(\mathbb{R})$.
It is worth noting that, patch-based approaches are only applicable on separable segmentation tasks.
\paragraph{Definition (Separability)} Using the previous notation, a segmentation task associated to $f$ is said to be 
\emph{p-separable} if there exists a family of functions $\{g_k: \mathcal{M}_p(\mathbb{R}) \to \mathcal{M}_p(\{0,1\})\}_{k=1}^P$ such that
\begin{equation}
\label{eq: sep}
 f(X) = \sum_{k=1}^P \nu_k(g_k(\varepsilon_k(X))), \quad \text{for all}\ \ X \in \mathcal{I} .
\end{equation}
In particular, the separation is said to be simple --which we will assume in the sequel-- when $g_k = g$ for all $k$.
In other words, by assuming the separability condition, one hypothesizes that the patches extracted by $\varepsilon_k$ 
contain enough information to classify each element therein. With this condition, the $d^2$-dimensional problem of segmenting an image is now broken
down into a series $p^2$-dimensional one. 
However, as aforementioned, the drawbacks of the patch-based works include: 
\begin{itemize}
 \item[-] \emph{The time complexity}: If we assume the patch extractors and mergers ($i.e.$ the $\varepsilon_k$'s and $\nu_k$'s) 
	  to have a $O(1)$ complexity then the overall segmentation will be in $O(PC_g)$ where $C_g$ is the time complexity of 
	  the patch model $g$ and $P$ is the number of patches. %Taking a small number of patches will reduce the complexity, 
	  %while introducing grid patterns in the final segmentation. 
	  The commonly used trick
	  parallelizes the procedure and is still time consuming for large values of $P$, as shown in our experiments.
 \item[-] \emph{The overall segmentation homogeneity}: On the other hand, in practice, the separability assumption leads to a lack
	  of overall segmentation homogeneity. That is, the patch model $g$ of Eq. \ref{eq: sep} is restricted on a small
	  neighborhood and may fail for some objects where a larger neighborhood is required for better results. 
	  Thus the need to an additional post-processing step in many patch-based.
 \item[-] \emph{Breaking the posterior distribution}: Given an input patch $x$ and its associated true segmentation $s$, 
	  the patch network $g$ approximate the conditional distribution $\mathbb{P}(s|x)$. However, after a standard patch 
	  aggregation procedure, the final segmentation is no longer following $g$'s distribution. 
	  This problem has been clearly discussed in the context of inverse imaging problems in \citet{Zoran_2011}. 
	  Therefore, it might be beneficial to learn complex aggregation models.
\end{itemize}

In this paper, the function $f$ is approximated by a FCNN function $\tilde{f}$, characterized by $\theta$, 
the set of weights and biases of Eq. \ref{eq: cnn}.  Our framework aims at avoiding any aggregation step and ensures an overall
segmentation homogeneity. To do so, we start by assuming the simple separability assumption of Eq. \ref{eq: sep} to train a patch network $\tilde{g}$
to obtain a well optimized set of weights $\theta_g$. Then, to decrease the running time and improve the overall homogeneity, 
we adapt $\theta_g$ to capture more information about the entire image, that is
the final network $\tilde{f}$ will have as parameters  $\theta_f$, obtained by fine-tuning $\theta_g$.
The procedures are explained step by step in the sequel and are depicted in Figure \ref{fig: fwj}.

\subsection{The Different Phases in Practice}
Each phase depicted in Figure \ref{fig: fwj} of the proposed framework is explained in the following.
\begin{itemize}
\item \textbf{Phase 1:} Patches are extracted from the training images to build the patch database. The latter
becomes the source database, denoted $\mathcal{D}_P$. Recall that it is not 
enough to extract a large number of patches. One should also ensure the diversity of the patches 
in order to have 
in a balanced proportion all the possible cases. The extraction can be made in a grid manner or randomly. 
\item \textbf{Phase 2:} A network $f_P$ is freely designed and trained, from scratch, on the 
previously built patch database $\mathcal{D}_P$.
With a well constructed patch
database, it becomes easier to train $f_P$ from scratch, since $\mathcal{D}_P$ provides the 
major requirements for training: enough training samples and diversification. Recall that,
to segment an original full-sized image with $f_P$ a patch extraction step is performed followed by patch 
segmentation and finally a patch aggregation step is needed.
\item \textbf{Phase 3:} The network $f_I$ is built based on the architecture of $f_P$. 
The transfer learning occurs here in the framework. In this work, the transfer consists in copying all the 
weights of $f_P$ into $f_I$. To assign a label (blood vessels, optic disc) to
a pixel, a small neighborhood is analyzed. The latter can be found both in a patch containing the pixel and 
(obviously) in the entire image. Thus, it is natural to infer that the weights used to segment patches should
produce acceptable results when applied on the original images.  
\item \textbf{Phase 4:} This final phase aims at adapting the transferred weights (of $f_P$) 
to work properly on $\mathcal{D}_I$. These weights, being learned from small patches, may not take
into account the overall homogeneity of a full-sized image segmentation. As a consequence, the 
network $f_I$ is fine-tuned on the full-sized images. In practice, this phase should be handled
with special attention to avoid a negative transfer and/or an over-fitting.
\end{itemize}
 
How general is the framework? Notice that commonly used training procedures can be cast as special cases of the proposed
framework. Literature training methods can be grouped in two categories ($see$ Section \ref{S:rw}): patch-based, images-based.
The patch-based methods are equivalent to using only the phases 1 and 2 of the framework. Recall that an additional patch aggregation
is needed to obtain the final segmentation.
%The images from scratch group can be seen as using only the phase 4 of the framework. 
By setting the patch database to be ImageNet and using the VGG network as $f_P$, 
habitual transfer learning based training propositions can be cast as special case of the proposed framework.

\begin{figure*}  
    \centering
    \centerline{\includegraphics[width=14cm]{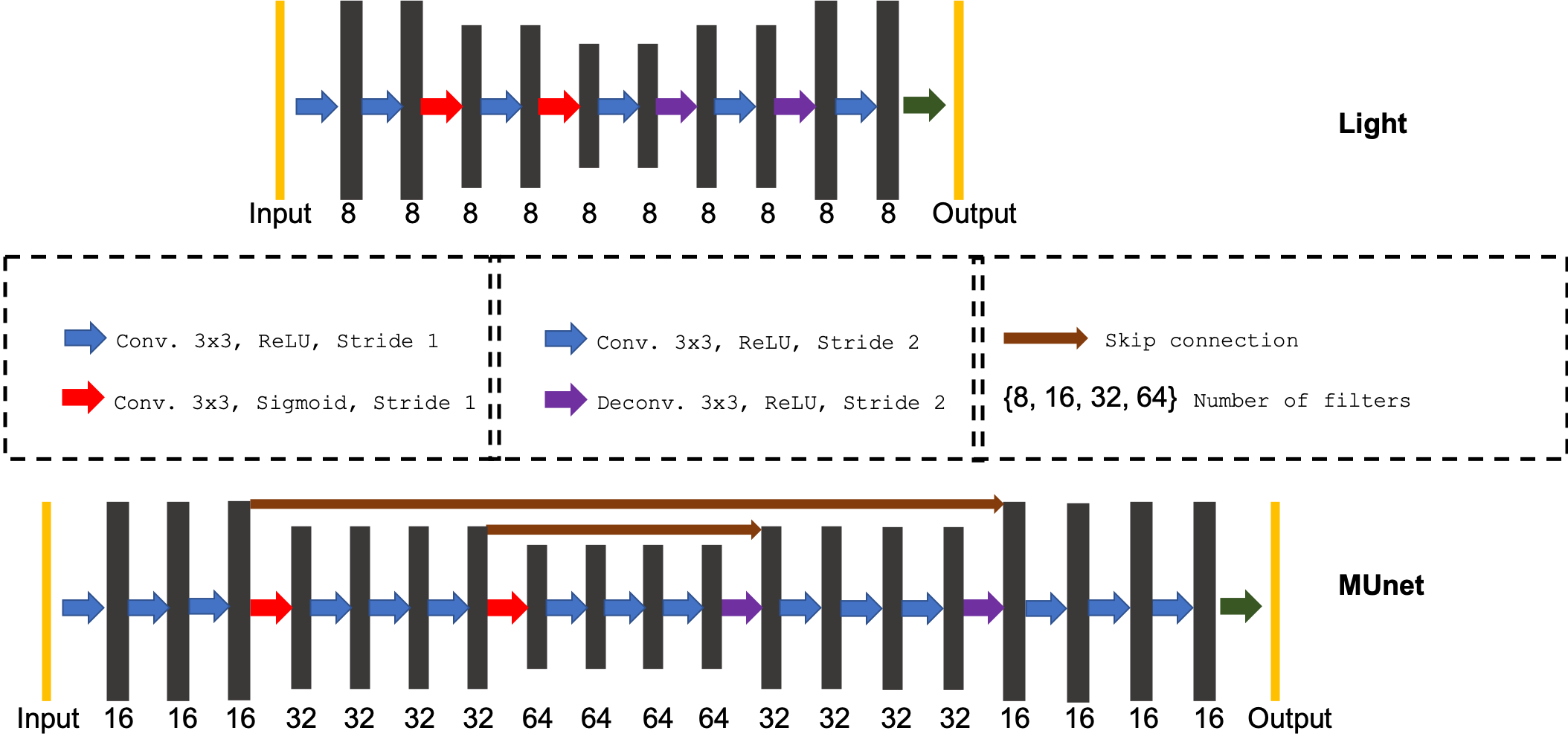}}
    \centerline{\includegraphics[width=14cm]{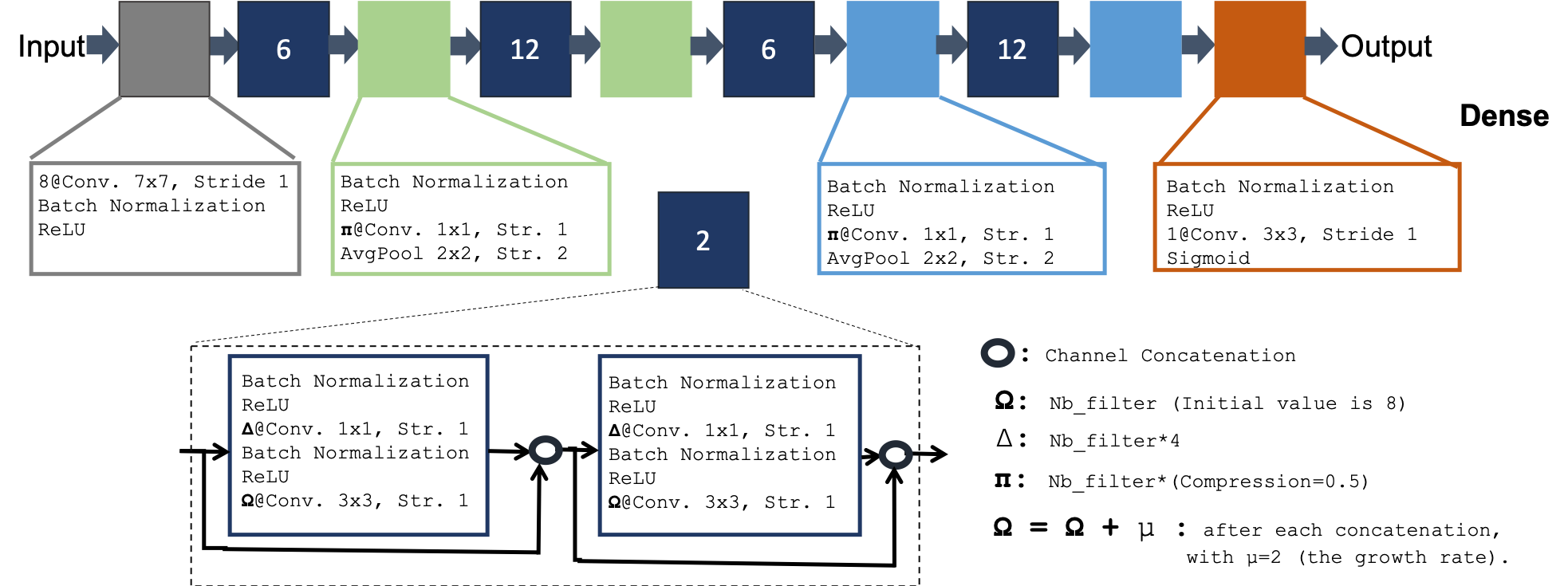}}
    \caption{ Illustration of the used networks. The layers used in Light and Mini-Unet are described in between their drawing.
    The ones in the Dense network are grouped by block. The dark-blue squares correspond to the densely connected parts and the
    inscribed digit within is the number of blocks (an example is given with $2$ dense blocks). 
    (For interpretation of the
    references to color in this figure legend, the reader is referred to the web version
    of this article.)}
    \label{fig: networks}  
\end{figure*}

\section{Experiments}
\label{S:ex} 
This section details all the experimental set-up and the final results, starting with a brief presentation
of the databases.

\subsection{Databases}
\label{S:ex:db} 
The following databases are considered:  DRIVE, STARE, CHASE\_DB1 for the task of RBVS and IDRiD for the one of ODS.
On the first three the test images are manually annotated by two
human observers, and as practiced in the literature, we used the first observer 
as a ground-truth to evaluate our methods. \rev{On all the databases, the images are resized
$(i)$ for computational reasons to avoid memory overload, and $(ii)$ because of the 
two down-sampling layers in the networks ($i.e.$ the input size should be a multiple of $4$)}.
However, all the segmented images are 
resized back to their initial size to compute the metrics, to have a fair comparison with 
the existing methods. In the following, the initial and resized sizes are given for each database.

\rev{On all the databases, a data augmentation is performed using commonly used  operations such as vertical and horizontal flipping, random translations, etc.}

\textbf{DRIVE} (Digital Retinal Images for Vessel Extraction) 
\bib{DRIVE}\footnote{\url{http://www.isi.uu.nl/Research/Databases/DRIVE/}} is a dataset
of 40 expert annotated color retinal images taken with a fundus camera of size $584 \times 565 \times 3$.
The database
is divided into two folds of 20 images: the training and testing sets. A mask image delineating
the field of view (FOV) is provided but not used in our experiments. The images are resized to 
$584 \times 568 \times 3$.

\textbf{STARE} 
(STructured Analysis of the REtina)\footnote{\url{http://www.ces.clemson.edu/~ahoover/stare/}}
\bib{STARE} is another well known, 
publicly available, database. The dataset is composed of 20 color fundus photographs captured with a
TopCon TRV-50 fundus camera  at $35^\circ$  field of view. Ten images are pathological cases 
and contain abnormalities that make the segmentation task even harder. Unlike DRIVE, 
this dataset does not come with a train/test split, thus a 5-fold cross-validation is performed. 
The images are of size $605\times 700$ and resized to $508 \times 600 \times 3$.

\textbf{CHASE\_DB1}\footnote{\url{https://blogs.kingston.ac.uk/retinal/chasedb1/}}  \bib{CHASE}
consists of  28 fundus images of size $960\times 999\times 3$. They are acquired  
from both the left and right eyes of 14 children. On this dataset, a 4-fold cross-validation is 
performed. The images are resized to $460\times 500 \times 3$.

\textbf{IDRiD} (Indian Diabetic Retinopathy Image Dataset)\footnote{\url{https://idrid.grand-challenge.org/}} \bib{IDRID}
is a publicly available dataset introduced during a 2018 ISBI challenge. It provides expert annotations of 
typical
diabetic retinopathy lesions and normal retinal structures. The images were captured with a kowa VX-10 
alpha  digital fundus camera. The optic disc segmentation task is considered in this paper.
IDRiD consists  of 54 training images of size $2848\times  4288\times 3$
and 41 testing ones. In our experiments, the images are downscaled by $10$ to obtain $284\times 428 \times 3$
size.

%\textbf{IOSTAR}\footnote{\url{http://www.retinacheck.org/datasets}} \bib{zhang_2016_IOSTAR} TBD

\begin{figure*}
    \centering
  \centerline{\includegraphics[width=11cm]{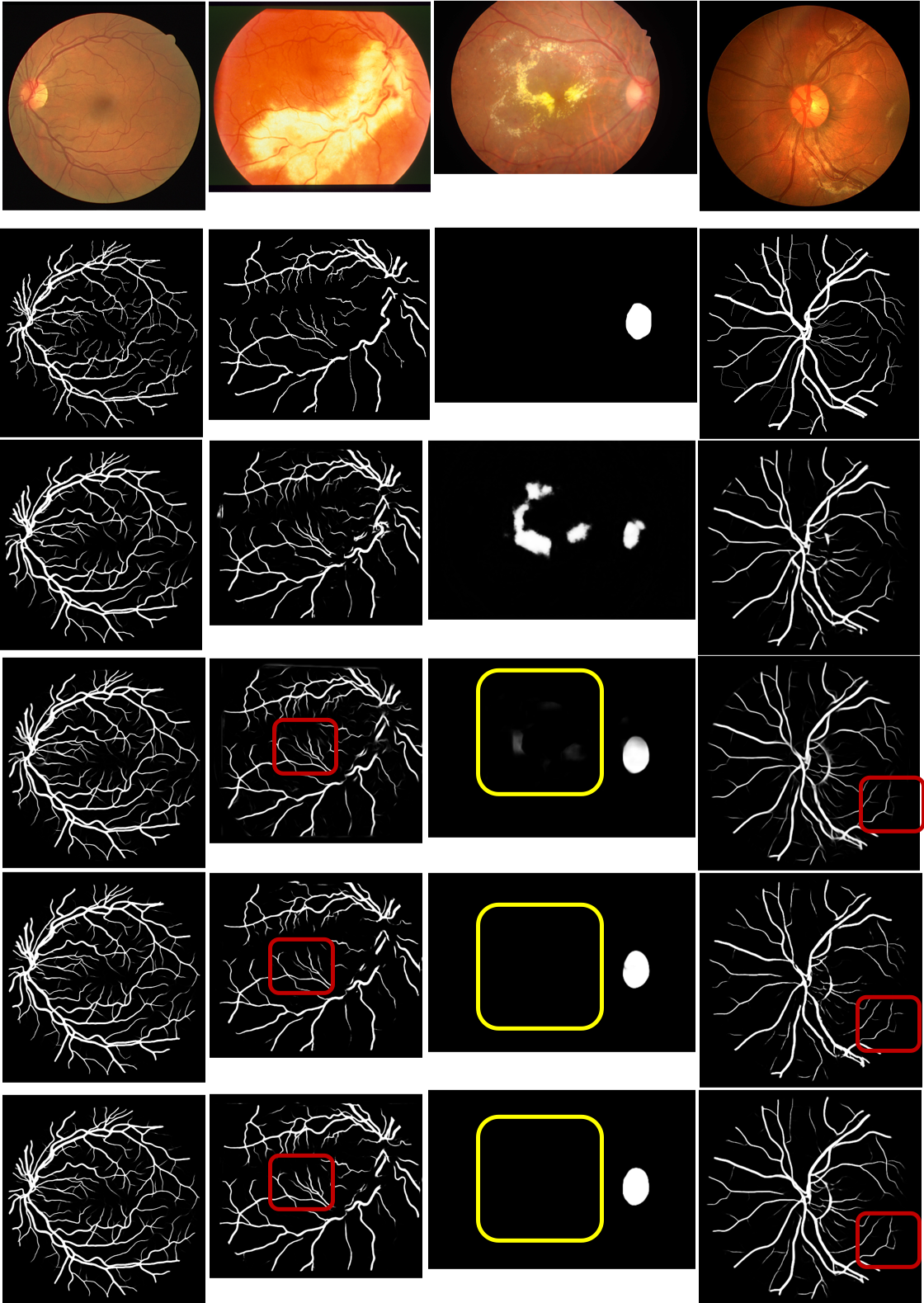}}
    \caption{Qualitative results using the \textbf{Dense} network on: the 01 DRIVE image, the 0044 STARE image, the 66 IDRiD image, and the 02L CHASE one. 
    From top to bottom: $(1)$ the original image, $(2)$ the ground-truth, $(3)$ the result after a training from scratch, $(4)$
    the patch-based segmentation, $(5)$ the result a the image-level with a frozen transfer, and
    $(6)$the result after fine-tuning.The red arrows show some cases where the probabilities are strengthen after
    fine-tuning to obtain solid lines while the yellow ones show an examples of false positives'
    removing. (For interpretation of the
    references to color in this figure legend, the reader is referred to the web version
    of this article.) }
    \label{fig: res_images}  
\end{figure*}

\subsubsection{Building the patch databases} 
The patches are extracted based on the distribution of the class labels in the ground-truths. On the task
of RBVS, as illustrated on Fig. \ref{fig: res_images}, the blood vessels are spread across the ground-truth. 
As a consequence, the patches are 
extracted with a stride of 16. While on IDRiD the optic disc pixels are localized in a small region of the 
ground-truth image. Hence, the patch database is built by extracting 500 patches centered on randomly 
selected optic disc pixel and another 500 on the background pixels. 

\subsubsection{Pre-processing} The success of DCNN-based models depends on the pre-processing performed on the data beforehand. The
pre-processing step aims mainly at improving the quality of the input data. The following pre-processing is performed 
on each RGB image of all the databases:
$(1)$ gray-scale conversion, $(2)$ a gamma correction (with gamma set to $1.7$), and $(3)$ a Contrast Limited Adaptive 
Histogram Equalization (CLAHE).  The patches on all databases are extracted after the pre-processing step.

\subsection{The Networks}
\label{S:ex:net}

Three networks have been used in this paper: a standard lightweight FCNN of $8,889$ weights (Light), a reduced version
of the U-Net of $316,657$ weights (Mini-Unet), and a densely connected network of $1,032,588$ parameters (Dense).  
All these networks have an end-to-end structure 
and can thus process inputs of arbitrarily sized ($i.e.$ patches or images). The architectural
details are depicted in the figure \ref{fig: networks}.

The \textbf{Light} network is a simple FCNN with $8$ filters on each layer. With its $8,889$ weights
this network is very light and very suitable for low memory embedded systems. Moreover, it also provides some insights
on the complexity of the task at hand. In other words, if the Light network's performances are not very far 
from the other (state-of-the-art) complex networks, then one can conclude that the task is not very difficult 
and can be solved with a small number of parameters.

The \textbf{Mini-Unet} network with  $316,657$ trainable parameters is a reduced version of the U-Net\bib{UNET}. It is 
less complex than the Dense network in terms of architecture and uses a relatively small number of weights
in each layer.

The \textbf{Dense} network consists of an overall of  $1,032,588$ trainable parameters. It is based on 
the densely connected design of \citet{Huang_2017_Densely}. In the dense blocks (dark-blue 
squares), the output of each intermediate layer $\ell_i$ is stack channel-wise on top of the one of $\ell_{i-1}$ to feed the layer $\ell_{i+1}$. In this network, three variables ($\Omega, \Pi$, and $ \Delta$) are used to adapt the number of filters after each layer. $\Omega$ is initially set to $8$ and increased by $2$ after each concatenation
while $\Delta = \Omega*4$ and $\Pi = \Omega*0.5$. 

\subsection{Training setup}

Regarding the training procedure, it has been carried out as described in the following.
\begin{itemize}
    \item A combined loss function which is defined, given a ground-truth $\mathbf{y} \in \{0,1\}^n$ and a prediction
     $\mathbf{p} \in [0,1]^n$, by 
     \[ \ell(\mathbf{y, p}) = - \langle\mathbf{y}, \log(\mathbf{p})\rangle 
      - \langle \mathbf{1} - \mathbf{y}, \log(\mathbf{1} - \mathbf{p})\rangle
             - \frac{2 \langle\mathbf{y, p}\rangle }{\langle\mathbf{y, 1}\rangle + \langle\mathbf{p, 1}\rangle } ,\]
    where $\log(\mathbf{p}) = [\log(\mathbf{p}[1]), ..., \log(\mathbf{p}[n])]$, $\mathbf{1} = [1, ...,1]$ and 
    $\langle\cdot, \cdot\rangle$ is the standard dot product. The first two terms of the equation correspond to the so-called  binary 
    cross-entropy (CE) while the last term is an approximation of the Dice coefficient. 
    The CE minimization aims at bringing the two vectors $\mathbf{y}$ and $\mathbf{p}$
    (seen as probability distributions) as close as possible, while adding the Dice criterion will 
    maximize their overlap.
    \item The learning rate is decayed to avoid over-fitting and to ameliorate the convergence. In other words,
    the learning rate is multiplied by a factor of $0.2$ after $5$ epochs without improvement on a 
    validation set. The minimum value is set to $10^{-5}$.
    \item An early stopping criterion is also set to avoid over-fitting. When working on patches, the 
    training is stopped after $30$ epochs without improvement on a validation set. While at the image 
    level, the training is carried out until 300 epochs. 
    \item On all databases a validation set is created to control the training processes by holding out 
    $20\%$ of training images. The patches being extracted after this separation there is no overlap between 
    the validation patches and the training ones.
    \item The \textit{AdaDelta} \bib{Zeiler_2012_ADADELTA} learning algorithm, which uses first-order 
    information to adapt its steps, is used. A mini-batch training is performed with $32$ batch size 
    at the patch-level and $1$ at the image one. The initial learning rate is set to $1$ by default.
    However when fine-tuning at the image-level, it is sometimes suitable to start with smaller values.
    For instance we started with $2.10^{-4}$ when fine-tuning on IDRiD full-sized images using the Dense network.
\end{itemize}

\begin{table*}
\caption{Numerical results on DRIVE.}
\label{tab: drive}
\begin{center}
 
\setlength{\tabcolsep}{0.2em}  
\begin{tabular}{|l|ll|c|c|c|c|c|c|c|}  
\cline{1-10}
 &\multicolumn{2}{c}{ \bf Methods} & \bf AUC & \bf Spec & \bf Sens & \bf Acc & \bf Dice &\bf Jaccard & \bf AUPRC \\
\hline   
\hline  
\parbox[t]{2mm}{\multirow{10}{*}{\rotatebox[origin=c]{90}{Patch-based}}} &Light & Patch & .9811&.9806&.7996&.9646&.7970&.6629 &.8861 \\
& Mini-Unet & Patch  & .9865&.9805&.8415&.9682&.8218&.6978 & .9096\\
& Dense & Patch  & $\mathbf{.9874}$&.9816&.8398&.9690&$\mathbf{ .8252}$ & $\mathbf{.7026}$ & $\mathbf{ .9136}$\\
\cline{2-10}
&\multicolumn{2}{l|}{\citet{Liskowski_2016_Segmenting}} &.9790& .9807& .7811& .9535 &- &-&- \\ 
&\multicolumn{2}{l|}{\citet{Li_2016_Cross}} &.9738 &.9816 &.7569 & .9527 & - & - &-\\ 
&\multicolumn{2}{l|}{\citet{Dasgupta_2017_Fully}} &.9744&  .9801& .7691& .9533 & - & -&- \\ 
&\multicolumn{2}{l|}{\citet{Yang_2017_Patch}} &.9792 & $\mathbf{ .9839}$ &.7811 & .9560 & - & -&- \\ 
&\multicolumn{2}{l|}{\citet{Jiang_2018_Retinal}} &.9680 &.9832 &.7121 & .9593 & - & - &-\\ 
&\multicolumn{2}{l|}{\citet{Birgui_2018_Retinal} }& .9801 & .9837 & .7665 & .9558 & - & -&-\\ 
&\multicolumn{2}{l|}{\citet{Oliveira_1018_Fully}} &.9821 &.9804 &.8039 & .9576 & - & - &-\\ 
\hline
\hline
\hline    
\parbox[t]{2mm}{\multirow{12}{*}{\rotatebox[origin=c]{90}{Image-based}}} 
&\multirow{4}{*}{Light} & Image/Frozen  & .9800&.9781&.8113&.9634&.7939&.6586 & .8798\\
&& Image/Fine-tuned  & .9808& .9772&.8187&.9632&.7949 &.6600 & .8822\\ 
&& Image/Scratch  & .9764 & .9791 & .7808 & .9616 & .7797 & .6393 & .8646 \\ %Mean time: 0.3339 
\cline{2-10}
&\multirow{4}{*}{Mini-Unet} & Image/Frozen  & .9864&.9778& $\mathbf{ .8586}$&.9673&.8204&.6957&.9097\\
&& Image/Fine-tuned  & .9866 & .9822 & .8342 & .9691 & .8246 & .7017 &.9114\\
&& Image/Scratch  & .9851 & .9813 & .8273 & .9676 & .8164 & .6900 & .9045 \\
\cline{2-10}
&\multirow{4}{*}{Dense} & Image/Frozen  & .9869&.9798&.8494&.9683&.8235&.7001 & .9112\\
&& Image/Fine-tuned  & .9870 & .9830 & .8277 & $\mathbf{ .9693}$ & .8242 & .7013 & .9129\\ 
&& Image/Scratch  & .9852 & .9834 & .8134 & .9684 & .8174 &.6914 & .9047 \\  
\cline{2-10}   
&\multicolumn{2}{l|}{\citet{Fu_2016}} &-&  -& .7294 & .9470 & - & - &-\\ 
&\multicolumn{2}{l|}{\citet{Maninis_2016}} & - & - & - & - & .8220 & - &-\\   
&\multicolumn{2}{l|}{\citet{Mo_2017_Multi}}   & .9782 & .9780 & .7779 & .9521 & -&-&-\\ 
&\multicolumn{2}{l|}{\citet{Birgui_2018_Retinal}}  & .9787 & .9782 &  .7990 & .9552 & -&- &-\\
\hline 
\end{tabular}  
\end{center}
\end{table*}  
  
\begin{table*}
\caption{Numerical results on STARE.}
\label{tab: stare}
\begin{center}
 
\setlength{\tabcolsep}{0.2em}  
\begin{tabular}{|l|ll|c|c|c|c|c|c|c|}  
\cline{1-10}
 &\multicolumn{2}{c}{ \bf Methods} & \bf AUC & \bf Spec & \bf Sens & \bf Acc & \bf Dice &\bf Jaccard & \bf AUPRC \\
\hline   
\hline  
\parbox[t]{2mm}{\multirow{5}{*}{\rotatebox[origin=c]{90}{Patch-based}}} 
&Light & Patch & .9780&.9833&.7259&.9643&.7456&.6000 & .8502 \\
& Mini-Unet & Patch  & .9854&.9861&.7731&.9702&.7914&.6579 & .8929\\
& Dense & Patch  & .9849 & $\mathbf{ .9868}$ & .7695 & .9705 & .7929 & .6682 &.8969 \\
\cline{2-10}  
&\multicolumn{2}{l|}{\citet{Liskowski_2016_Segmenting}} &$\mathbf{ .9928}$ & .9862 & .8554 & 
$\mathbf{.9729}$ &- &-&- \\
&\multicolumn{2}{l|}{\citet{Li_2016_Cross}} &.9879&  .9844 & .7726 & .9628 & - & - &-\\ 
&\multicolumn{2}{l|}{\citet{Jiang_2018_Retinal}} &.9870 &.9798 &.7820 & .9653 & - & -&- \\  
&\multicolumn{2}{l|}{\citet{Hajabdollahi_2018}} &- &.9757 &.7599 & .9581 & - & - &-\\ 
&\multicolumn{2}{l|}{\citet{Oliveira_1018_Fully}} &.9905 &.9858 &$\mathbf{.8315}$ & .9694 & - & -&- \\  
\hline
\hline
\hline      
\parbox[t]{2mm}{\multirow{8}{*}{\rotatebox[origin=c]{90}{Image-based}}} 
&\multirow{4}{*}{Light} & Image/Frozen & .9743&.9800 & .7379&.9619 & .7405& .5913 & .8268\\
&& Image/Fine-tuned & .9777 & .9813 &  .7434 & .9636 & .7499 & .6037 & .8424 \\
&& Image/Scratch & .9714 & .9762 & .7257 & .9576 & .7181 & .5642 & .8067\\  
\cline{2-10}
&\multirow{4}{*}{Mini-Unet} & Image/Frozen  & .9817&.9844&.7858&.9694&.7913&.6570 & .8757\\
&& Image/Fine-tuned & .9867 & .9865 & .7955 & .9723 & .8083 & .6805 & .8966 \\ 
&& Image/Scratch   & .9856 & .9842  & .7956 & .9699 & .7993 & .6676 & .8897  \\ 
\cline{2-10}
&\multirow{4}{*}{Dense} & Image/Frozen  & .9831 & .9852 & .7907 & .9704 & .7991 & .6607 & .8782 \\ 
&& Image/Fine-tuned & .9869&.9844&.8152&.9718&.8100& $\mathbf{ .6827}$ &  $\mathbf{.9033}$\\
&& Image/Scratch  &  .9836&.9854&.7792&.9702&.7915&.6588 & .8863\\ 
\cline{2-10}    
&\multicolumn{2}{l|}{\citet{Fu_2016}} &-&  -& .7140 & .9545 & - & - &-\\ 
&\multicolumn{2}{l|}{\citet{Maninis_2016}} & - & - & - & - & $\mathbf{ .8310}$ & -&- \\ 
&\multicolumn{2}{l|}{\citet{Mo_2017_Multi}}   &  .9885 & .9844 & .8147 & .9674  & -&-&-\\   
\hline 
\end{tabular}  
\end{center}
\end{table*}  

\subsection{Results} 
Some Qualitative results obtained using the Dense network are illustrated in the figures \ref{fig: res_images}\rev{, \ref{fig: res_images_munet}, and \ref{fig: res_images_light}}. 
The quantitative results are depicted in the tables \ref{tab: drive}, \ref{tab: stare}, \ref{tab: chase},
and \ref{tab: idrid}. 
For each proposition line the following nomenclature is used: \textbf{``NetName InputType/TransferType''},
where 
NetName $\in$ \{\textit{Light}, \textit{Mini-Unet}, \textit{Dense}\}, InputType $\in$ \{\textit{Image}, \textit{Patch}\}
and TransferType $\in $ \{\textit{Frozen}, \textit{Fine-tuned}, \textit{Scratch}\}. TransferType is only used when
InputType is $Image$. When the image network $f_I$ is trained from
a random initialization, TransferType is set to $Scratch$. Whereas $Frozen$ means the metrics are computed just after 
the initialization of $f_I$ with the weights of $f_P$, and $Fine-tuned$ is used when $f_I$
has been fine-tuned on $\mathcal{D}_I$.

On all the tables the methods
are grouped by their input type (patch- and image-based).
Five metrics are initially used to evaluate the methods: the Area Under the receiver operating characteristic Curve (\textbf{AUC}), 
computed with the scikit-learn\footnote{\url{http://scikit-learn.org/stable/index.html}} library, the sensitivity (\textbf{Sens}), the specificity (\textbf{Spec}), and the accuracy (\textbf{Acc})
computed after a $0.5$ threshold on the predictions using the following: 
\[\mathbf{Sens} = \frac{TP}{TP+FN}, \ \mathbf{Spec}  = \frac{TN}{TN+FP}, \]
\[\mathbf{Acc} = \frac{TP + TN}{TP+TN+FP+FN}, \]
where  $TP$, $TN$, $FP$, and $FN$  denote respectively the number of true positives, true negatives, false positives, and false negatives.
Additionally, the Dice and Jaccard coefficient are also considered to evaluate the overlap 
between the ground-truths and the predictions. They are given by
\[\mathbf{Dice} = \frac{2|G \cap P|}{ |G| + |P|}, \quad \mathbf{Jaccard} = \frac{|G \cap P|}{ |G \cup P|},  \]
where $G$ is the set of ground-truth pixels, $P$ is the prediction one, and $|X|$ is the number of elements in the set $X$. \rev{The Area Under the Precision/Recall Curve (\textbf{AUPRC}) is also computed as a complementary metric, using 
the scikit-learn library. }

\rev{The figures \ref{fig: IDRiD_Jaccard_box}, \ref{fig: CHASE_Jaccard_box}, 
\ref{fig: STARE_Jaccard_box},
and \ref{fig: DRIVE_Jaccard_box} provides the Jaccard boxplots on respectively IDRiD, CHASE, STARE,
and DRIVE. The x-axis' of those figures is labeled as follows: 
 \textbf{``N/I/T''},
where 
\textbf{N} $\in$ \{\textit{L}, \textit{M}, \textit{D}\} is 
a short for Light, Mini-Unet or Dense, \textbf{I} $\in$ \{\textit{IMG}, \textit{PCH}\} refers
to image or patch,
and \textbf{T} $\in $ \{\textit{Fr}, \textit{Fi}, \textit{Scr}\} stands for frozen, fine-tuned
or scratch. }

Furthermore, we computed the segmentation time of each network on the DRIVE database both at the patch and image-level.
Table \ref{tab: time} presents the segmentation time, in seconds, computed from a 
Tesla P100-PCIE-16GB GPU and an Intel i7-6700HQ CPU (with 8 cores of 2.60GHz and 16GB RAM). The
networks are implemented using Keras/TensorFlow\footnote{\url{https://www.tensorflow.org/}}. 
Note that the patches are segmented in a
multiprocessing setting using the Keras library.
    
\begin{table*}
\caption{Numerical results on CHASE-DB1.}
\label{tab: chase}
\begin{center}
 
\setlength{\tabcolsep}{0.2em}  
\begin{tabular}{|l|ll|c|c|c|c|c|c|c|}  
\cline{1-10}
 &\multicolumn{2}{c}{ \bf Methods} & \bf AUC & \bf Spec & \bf Sens & \bf Acc & \bf Dice &\bf Jaccard & \bf AUPRC \\
\hline   
\hline  
\parbox[t]{2mm}{\multirow{5}{*}{\rotatebox[origin=c]{90}{Patch-based}}} 
& Light & Patch  & .9775&.9834&.7350&.9660&.7490&.5997 & .8332 \\
& Mini-Unet & Patch  & .9849&.9867&.7761&.9719&.7916&.6558 &.8785 \\
& Dense & Patch  &  .9863& $\mathbf{ .9888}$&.7556&.9723&.7892&.6524 & .8851 \\
\cline{2-10}  
&\multicolumn{2}{l|}{\citet{Liskowski_2016_Segmenting}} & .9845 & .9668& $\mathbf{.8793}$ & .9577 & - &- & - \\ 
&\multicolumn{2}{l|}{\citet{Li_2016_Cross}} &.9716&  .9793 & .7507 & .9581 & - & - & - \\ 
&\multicolumn{2}{l|}{\citet{Jiang_2018_Retinal}} &.9580 &.9770 &.7217 & .9591 & - & -&- \\ 
&\multicolumn{2}{l|}{\citet{Oliveira_1018_Fully}} &.9855 &.9864 &.7779 & .9653 & - & -&- \\ 
\hline
\hline
\hline      
\parbox[t]{2mm}{\multirow{8}{*}{\rotatebox[origin=c]{90}{Image-based}}} 
&\multirow{4}{*}{Light} & Image/Frozen & .9751&.9797&.7498&.9636&.7403&.5888 & .8126\\
&& Image/Fine-tuned & .9780 & .9819 & .7535 & .9659&.7530 & .6049 & .8307\\
&& Image/Scratch  & .9743 & .9790 & .7444 & .9625 & .7330 & .5793 & .8063 \\  
\cline{2-10}
&\multirow{4}{*}{Mini-Unet} & Image/Frozen  & .9837&.9829&.8072&.9705&.7898&.6535 & .8678 \\ 
&& Image/Fine-tuned & .9875&.9861&.8117&.9737& $\mathbf{  .8093}$&  $\mathbf{.6804}$ & .8923\\
&& Image/Scratch  & .9876 & .9851 & .8158 & .9733 & .8080 & .6785 & .8897 \\ 
\cline{2-10}
&\multirow{4}{*}{Dense} & Image/Frozen  & .9834&.9868&.7653&.9711&.7839&.6453 & .8708\\
&& Image/Fine-tuned & $\mathbf{  .9878}$&.9869&.8016&  $\mathbf{ .9737}$&.8068&.6768 & $\mathbf{.8935}$\\
&& Image/Scratch   & .9857&.9861&.7968&.9726&.7999&.6671 & .8849\\
\cline{2-10}    
&\multicolumn{2}{l|}{\citet{Mo_2017_Multi}}   & .9812 & .9816 & .7661 & .9599 & -&-& -\\   
\hline 
\end{tabular}  
\end{center}
\end{table*}

\begin{table}[t]
\caption{Numerical results on IDRiD.}
\label{tab: idrid}
\begin{center}
 
\centering 
\setlength{\tabcolsep}{0.5em}   
\begin{tabular}{|l|ll|c|}  
\cline{1-4}
 &\multicolumn{2}{c}{ \bf Methods} &  \bf Jaccard \\
\hline   
\hline 
\parbox[t]{2mm}{\multirow{2}{*}{\rotatebox[origin=c]{90}{P-b}}} &Light & Patch & .6413  \\
& Mini-Unet & Patch  & .9035 \\
& Dense & Patch  & .9233 \\
\hline
\hline
\hline      
\parbox[t]{2mm}{\multirow{8}{*}{\rotatebox[origin=c]{90}{Image-based}}}
&\multirow{4}{*}{Light} & Image/Frozen  & .5463 \\ 
&& Image/Fine-tuned  & .7898  \\ 
&& Image/Scratch  & .7498 \\  
\cline{2-4}
&\multirow{4}{*}{Mini-Unet} & Image/Frozen  & .8551 \\ 
&& Image/Fine-tuned  & .9216  \\ 
&& Image/Scratch  & .8720 \\  
\cline{2-4}
&\multirow{4}{*}{Dense} & Image/Frozen  & .9337 \\ 
&& Image/Fine-tuned  & $\mathbf{ .9390 }$ \\ 
&& Image/Scratch  & .7678 \\  
\cline{2-4} 
&\multicolumn{2}{|l|}{Leaderbord (Mask RCNN \bib{He_MaskRCNN})}&  .9338 \\
\hline
\end{tabular}  
\end{center}
\end{table}

\subsection{Discussion}
\begin{figure*}
    \centering
  \centerline{\includegraphics[width=11cm]{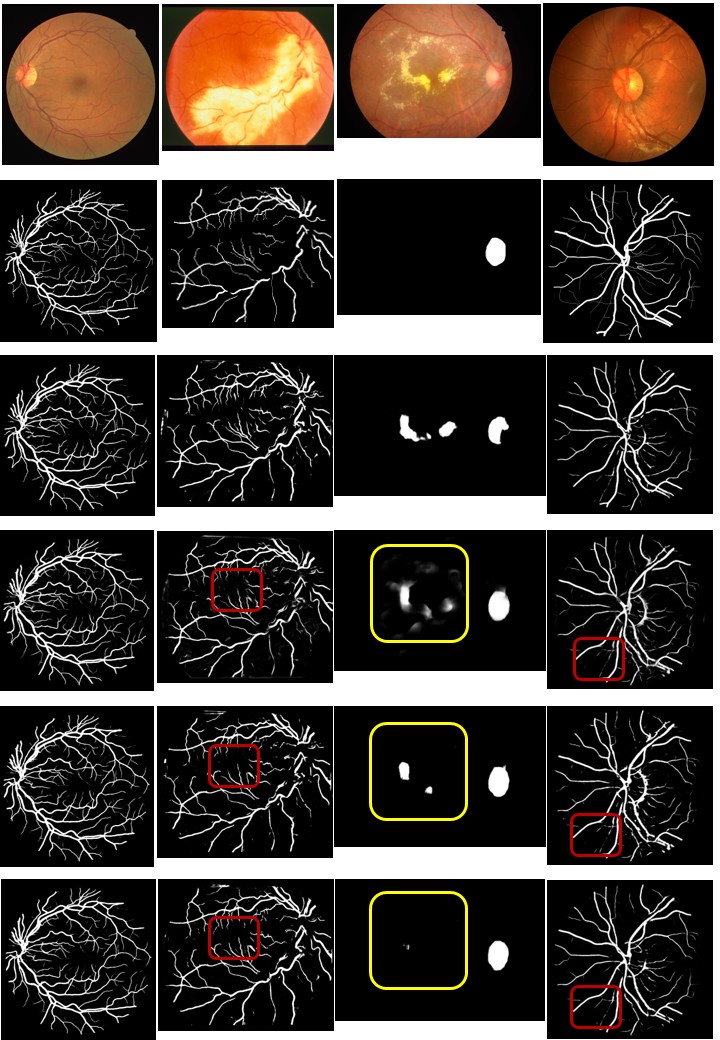}}
    \caption{\rev{Qualitative results using the \textbf{Mini-Unet} network on: the 01 DRIVE image, the 0044 STARE image, the 66 IDRiD image, and the 02L CHASE one. 
    From top to bottom: $(1)$ the original image, $(2)$ the ground-truth, $(3)$ the result after a training from scratch, $(4)$
    the patch-based segmentation, $(5)$ the result a the image-level with a frozen transfer, and
    $(6)$the result after fine-tuning. The red arrows show some cases where the probabilities are strengthen after
    fine-tuning to obtain solid lines while the yellow ones show an examples of false positives'
    removing. (For interpretation of the
    references to color in this figure legend, the reader is referred to the web version
    of this article.)} }
    \label{fig: res_images_munet}  
\end{figure*}

\begin{figure*}
    \centering
  \centerline{\includegraphics[width=11cm]{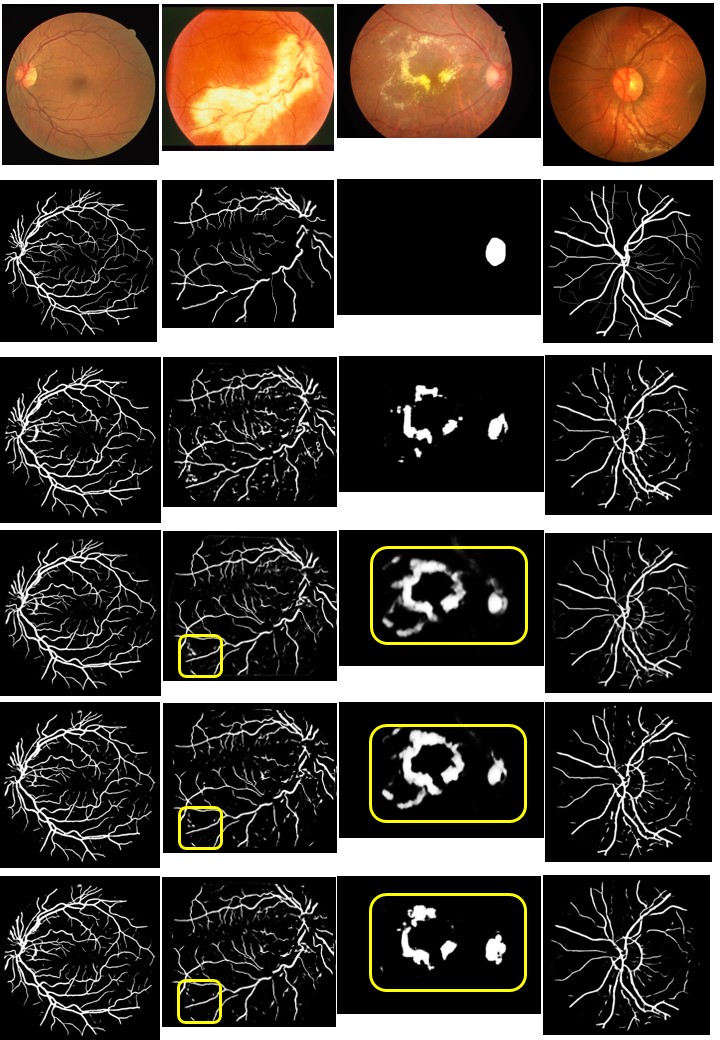}}
    \caption{\rev{Qualitative results using the \textbf{Light} network on: the 01 DRIVE image, the 0044 STARE image, the 66 IDRiD image, and the 02L CHASE one. 
    From top to bottom: $(1)$ the original image, $(2)$ the ground-truth, $(3)$ the result after a training from scratch, $(4)$
    the patch-based segmentation, $(5)$ the result a the image-level with a frozen transfer, and
    $(6)$the result after fine-tuning. The yellow boxes illustrates examples of false positives'
    removing. (For interpretation of the
    references to color in this figure legend, the reader is referred to the web version
    of this article.)} }
    \label{fig: res_images_light}  
\end{figure*}

As stated in the introduction, the principal question dealt with is: \emph{How to obtain state-of-the-art performances at the image-level using a freely designed network on the task of RIS?}
To answer that question the proposed  framework
starts by training the given network from the patches, then transfers the learned weights at the image-level.
The networks being freely designed and performing efficiently on the full-sized images, the proposed framework 
seems to correctly answer the principal question. 
\paragraph{Analysing the numerical results} On \textbf{DRIVE}, all the patch-based networks are improved 
after fine-tuning in terms of accuracy (Acc). The AUC is slightly reduced after fine-tuning on
the Light and Dense networks. However, the Dice and Jaccard of the patch-based Dense network 
are preserved after fine-tuning. Whereas, the fine-tuning of the Mini-Unet improved the patch metrics
in general. On \textbf{STARE}, all patch-based networks are improved in terms of Dice and Jaccard,
with $2\%$ Jaccard increase on Mini-Unet and Dense. Similarly, on \textbf{CHASE-DB1} and \textbf{IDRiD},
the fine-tuning improves the performances on all networks with $2\%$ Jaccard increase on Mini-Unet
and Dense. The Dense network being the best both at the patch-level and image one. 
\textbf{Overall}, the Dense network may be the best but the Light provides remarkable results 
despite its number of parameters except on the IDRiD dataset. 
\paragraph{Qualitative remarks}  As depicted in the figure \ref{fig: res_images}, the fine-tuning strengthen the 
class probabilities and reduces some false positives along the way. When the patch network achieves remarkable results,
with a careful fine-tuning, the qualitative results at the image-level will remain the same, as one can
observe on 
the DRIVE image of Figure \ref{fig: res_images} (on the left). 
\rev{The images selected varies from
simple to very complicated to segment: $(1)$ a simple image from DRIVE, $(2)$ one of the
most complicated cases of STARE, $(3)$ one of the most complicated cases of IDRiD, and
$(4)$ a medium case of CHASE\_DB1. Figures \ref{fig: res_images_munet} and 
\ref{fig: res_images_light} show respectively the predictions of the Mini-Unet and Light networks.}

\paragraph{How to set the initial learning rate?} This is still an open question in the field of deep learning. Hence,
the values used in this paper are mainly based on the preliminary work \bib{Birgui_2018_Retinal}.
For instance, when working at the patch-level, one should allow large learning rates
in the beginning to move as quickly as possible towards a local minimum but reducing it in time to stabilize 
the convergence. Whereas, when fine-tuning at the image-level, the image network is already well initialized, thus
the initial learning rate should be small. Consequently, our fine-tuning on the IDRiD images, using the Dense network, 
starts with a learning rate
of $2.10^{-4}$ and is decayed using the aforementioned criterion. However, on the other cases, we noticed that
an initial learning rate of $1$ leads to faster convergence in the fine-tuning.

\begin{figure*}  
    \centering
    \centerline{\includegraphics[width=7cm]{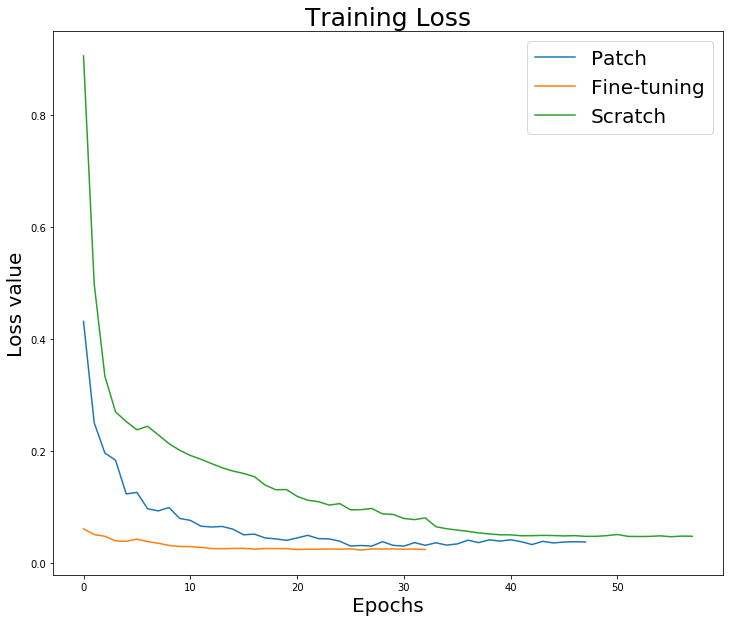}}
    \centerline{\includegraphics[width=7cm]{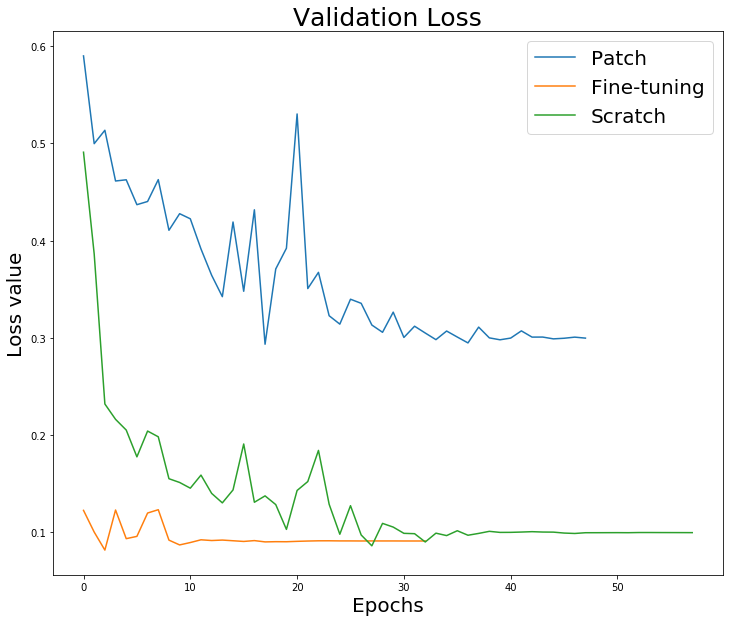}}
    \centerline{\includegraphics[width=7cm]{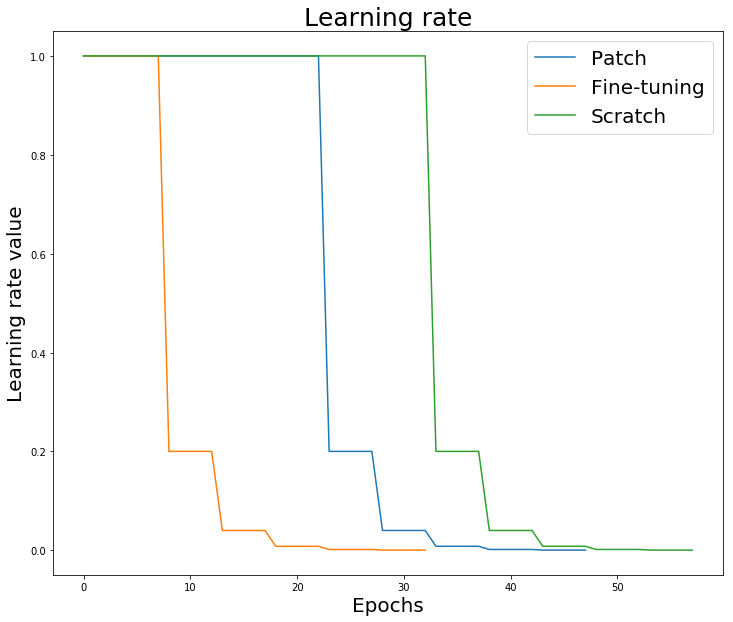}}
    \caption{ \rev{Illustration of the training curves of the Mini-Unet on the IDRiD database.
    (For interpretation of the
    references to color in this figure legend, the reader is referred to the web version
    of this article.)}}
    \label{fig: training_curves}  
\end{figure*}

\paragraph{The training curves}\rev{ Figure \ref{fig: training_curves} illustrates the training
curves of the Mini-Unet on the IDRiD database. The First row, which depicts the training loss
curves, shows the quality of patch training since the loss when fine-tuning starts very low 
and converges fast (less than 40 epochs). The training being stopped after 30 epochs without 
improvement on the validation set, the second row of Figure \ref{fig: training_curves}
shows the convergence in validation. Therefore the earlier stopping criterion occurs, in 
this case, after convergence on the validation set. The bottom row of Figure  
\ref{fig: training_curves} presents the learning rate values at each epoch.}

\paragraph{Analysing the framework's phases} The efficiency of the first phase is judged by the metrics of the 
second one. Indeed, with our patch extraction strategy, we trained efficiently three types of networks to reach
competitive results compared to other deep learning methods. Furthermore, it can be noticed that all our
patch-based metrics are either improved or kept after fine-tuning at the image-level. Therefore, the transfer learning (Phase 3)
and the fine-tuning (Phase 4) are not leading to a negative transfer. On the other hand, at the image-level, 
we noticed that it is always better to fine-tune the weights than training from scratch. This points out the importance 
of the Phase 1 and 2, that is a patch pre-training provides a better initialization point than a random one.
Also, the results show that it is better to fine-tune the weights at the image-level that keeping them frozen.
\paragraph{Comparing the RBVS methods} The comparison of the
retinal blood vessel segmentation methods is somehow difficult 
in the sense that $(i)$ some papers compute their results only in a restricted field of view, and $(ii)$ 
sometimes the train/test split may differ from one paper to another (when the database does not come with an explicit
train/test split). For instance, the proposition of \citet{Liskowski_2016_Segmenting} is a patch-based technique 
that only considers patches that are fully in the field of view. Whereas the results of \citet{Maninis_2016} on the
STARE database is based on a single random split of 10 images for training and the other 10 for testing, while the
other methods perform either a leave-one-out cross-validation or a 5-fold one. 
\rev{Therefore, the reported values may not be directly comparable in some cases}.
However, in all cases our framework can open the door for more research on freely designed networks and their application 
on embedded real-time systems. For instance, the proposed Light network seems to perform quite good compared to other
more complex and heavy networks and is suitable for some embedded systems with small memory. Compared to the 
quantization technique used by \bib{Hajabdollahi_2018} to obtain a low complexity network, our training strategy
applied to the Light network is performing better.
Moreover, we added the Jaccard coefficient which seems to be more suitable together 
with the Dice, since they provide insights on how perfect is the overlap of the predictions and the ground-truths. 
\paragraph{Comparing the ODS methods} The IDRiD database comes with an explicit train/test 
split which  facilitate the methods comparisons. Only the Jaccard is used as metric in the original challenge.
The baseline provided on the table \ref{tab: idrid} is the best result  from the leaderboard of the challenge
which is based on the state-of-the-art network for object localization (Mask-RCNN) of \citet{He_MaskRCNN}.
The Dense network provides the best result after fine-tuning. On the other hand, one can notice that the Light network
seems to struggle, thus one can conclude on the hardness of the task on IDRiD.
\paragraph{Running time} As depicted in Table \ref{tab: time}, one can see the importance of the image-based 
segmentation in terms of segmentation time. Indeed, even with a parallelized patch procedure, all the networks
are considerably faster at the image-level than at the patch one both on the GPU and the CPU.

\paragraph{Fine-tuned $v.s.$ Frozen transfer} \rev{It can be noticed that on simple databases such
as DRIVE the fine-tuned and frozen weights provide similar metrics. That is because 
$(a)$ the segmentation is relatively simple, and $(b)$ the patches contain all the necessary 
discriminative information. However, the difference between fine-tuned and frozen results 
becomes more obvious as the task gets complicated ($see$ the DICE metrics in tables \ref{tab: stare} 
and \ref{tab: chase}). On IDRiD (Table \ref{tab: idrid}), it is clear that when using simple
networks (Light and Mini-Unet) the fine-tuning improves considerably the results. While, the
Dense network happens to have enough complexity to capture the necessary underlying patterns
from the patch database and the fine-tuning does not have a large impact. In any cases, 
the fine-tuning is an important phase even though its impact may not always be large.
}

\begin{figure*}
    \centering
  \centerline{\includegraphics[width=16cm]{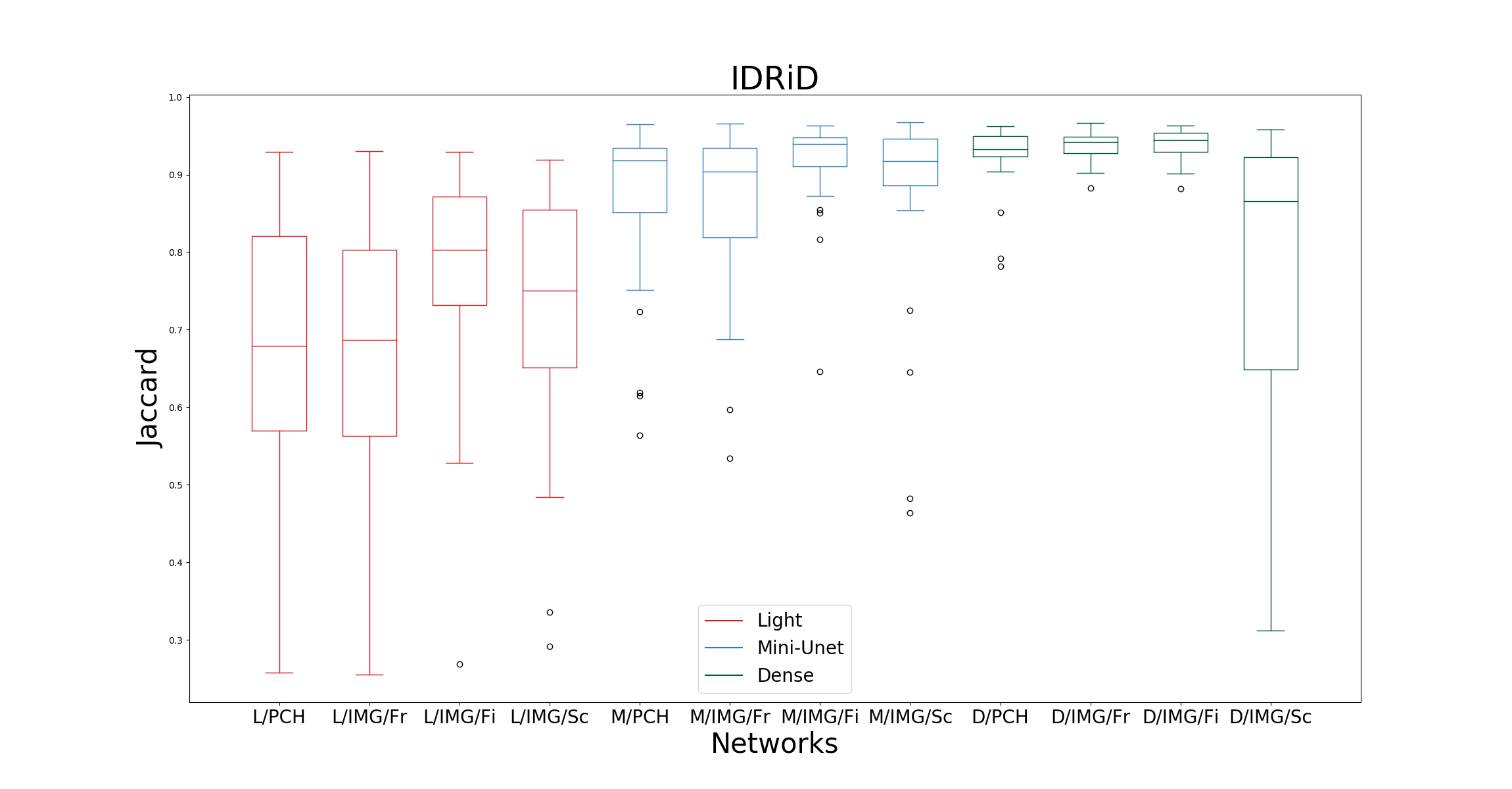}}
    \caption{\rev{Jaccard boxplot on IDRiD. (For interpretation of the
    references to color in this figure legend, the reader is referred to the web version
    of this article.)} }
    \label{fig: IDRiD_Jaccard_box}  
\end{figure*}

\begin{figure*}
    \centering
  \centerline{\includegraphics[width=16cm]{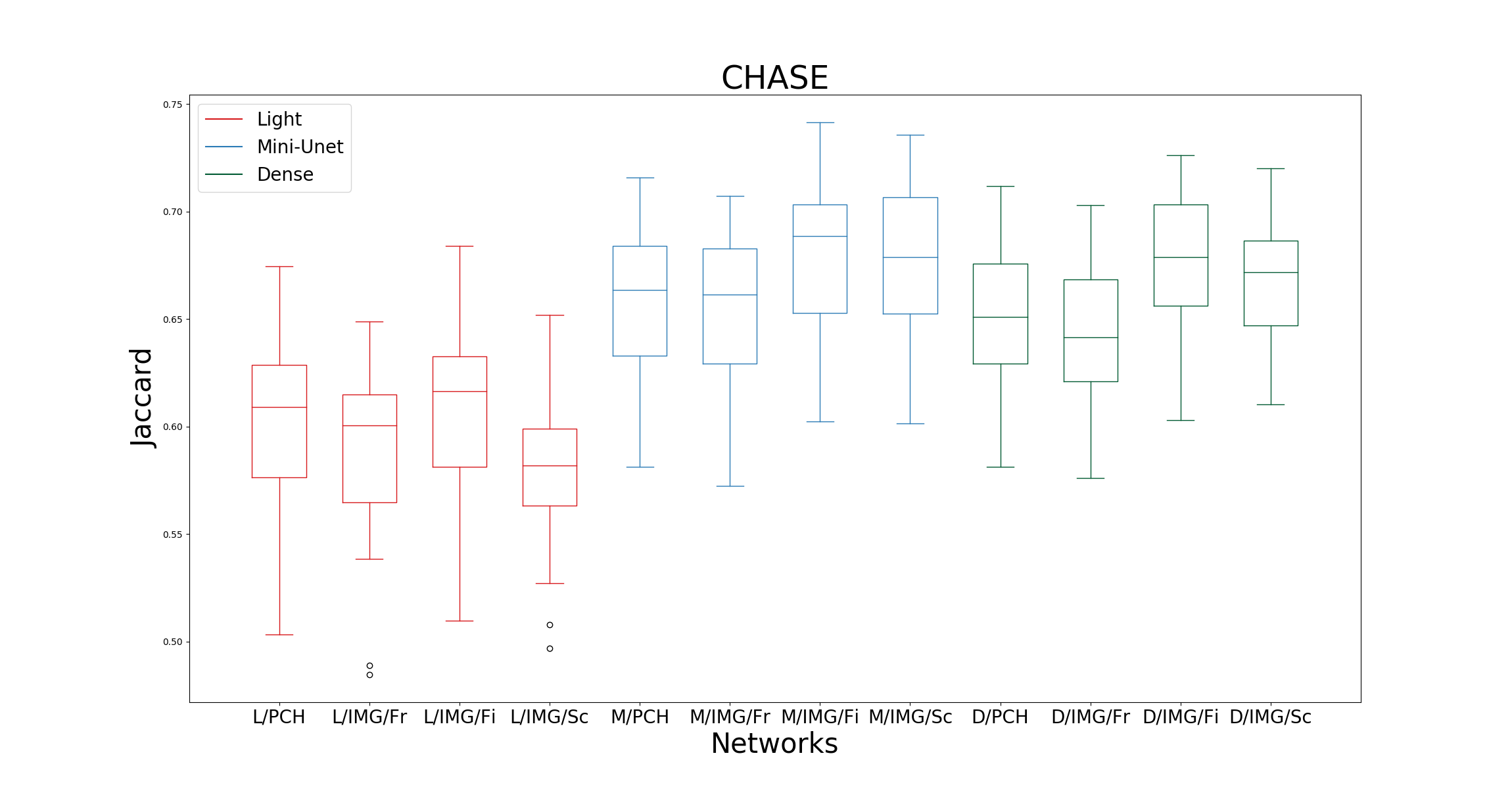}}
    \caption{\rev{Jaccard boxplot on CHASE\_DB1. (For interpretation of the
    references to color in this figure legend, the reader is referred to the web version
    of this article.)} }
    \label{fig: CHASE_Jaccard_box}  
\end{figure*}

\begin{figure*}
    \centering
  \centerline{\includegraphics[width=16cm]{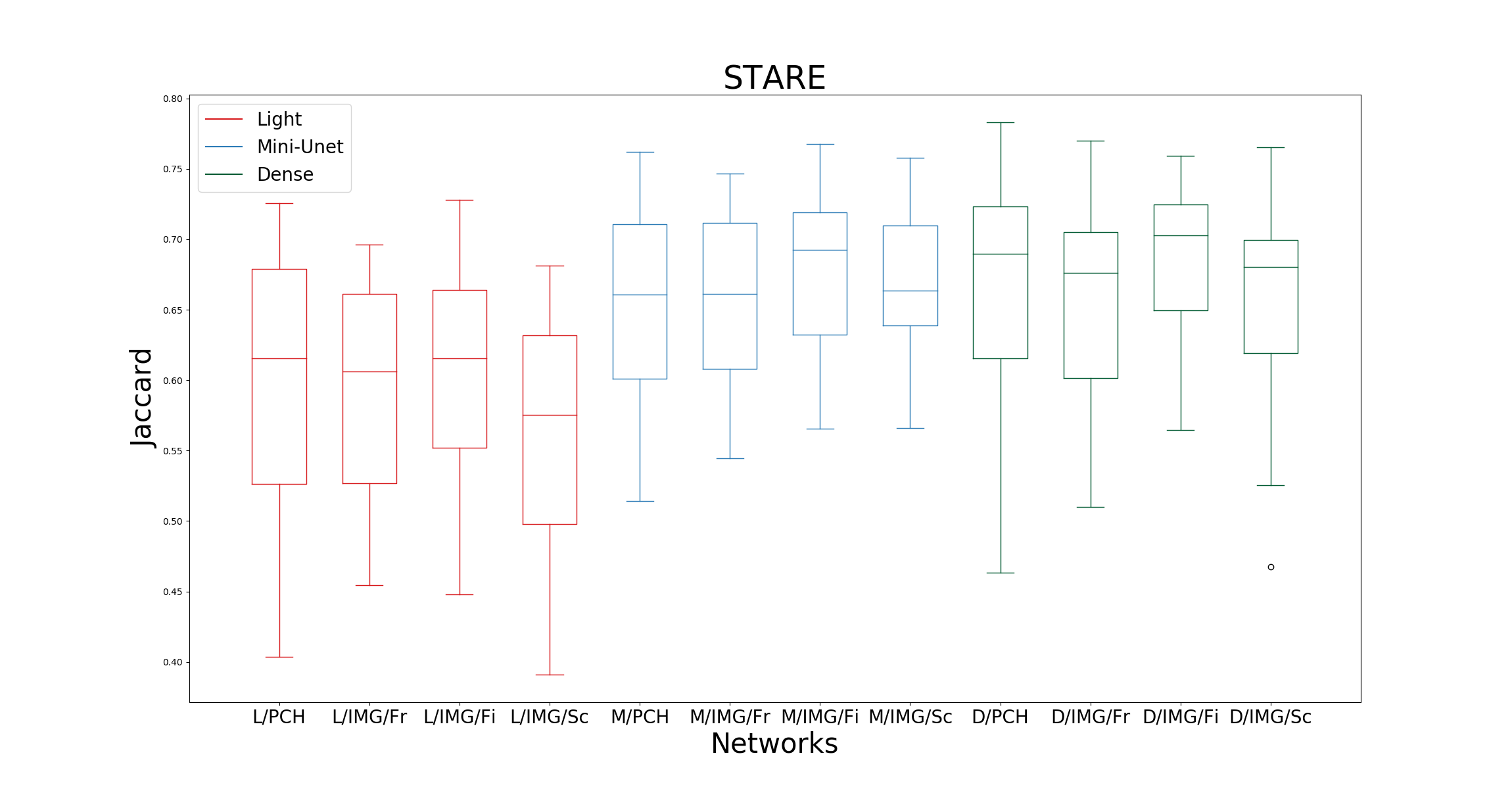}}
    \caption{\rev{Jaccard boxplot on STARE. (For interpretation of the
    references to color in this figure legend, the reader is referred to the web version
    of this article.)} }
    \label{fig: STARE_Jaccard_box}  
\end{figure*}

\begin{figure*}
    \centering
  \centerline{\includegraphics[width=16cm]{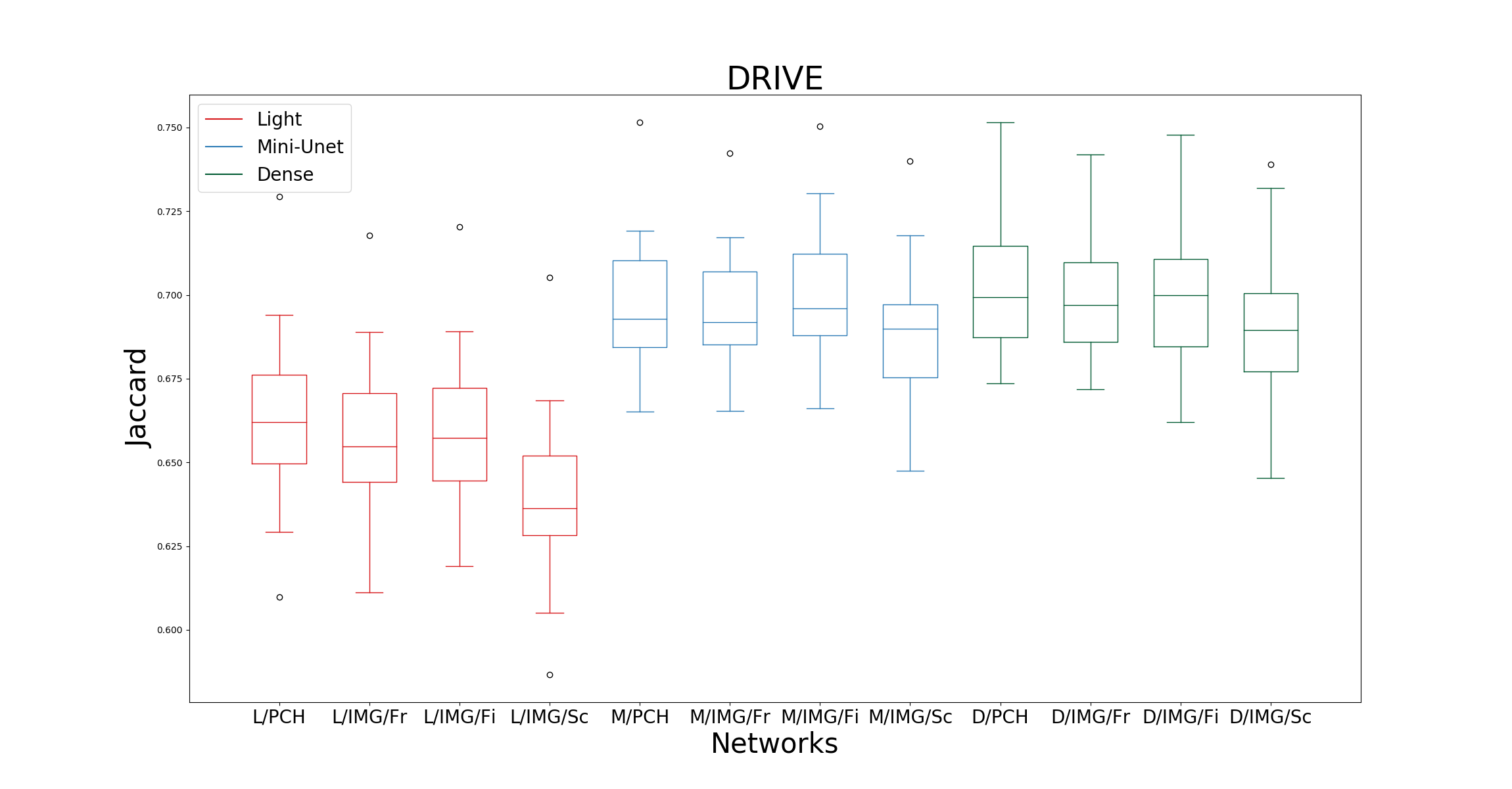}}
    \caption{\rev{Jaccard boxplot on DRIVE. (For interpretation of the
    references to color in this figure legend, the reader is referred to the web version
    of this article.)} }
    \label{fig: DRIVE_Jaccard_box}  
\end{figure*}

\paragraph{The boxplot analysis} \rev{The figures \ref{fig: IDRiD_Jaccard_box}, 
\ref{fig: CHASE_Jaccard_box}, \ref{fig: STARE_Jaccard_box}, and \ref{fig: DRIVE_Jaccard_box} 
provides the Jaccard boxplots on respectively IDRiD, CHASE, STARE, and DRIVE.  On the
task of RBVS, the fine-tuned Light network outperforms many state-of-the-art methods, 
looking at the AUC, Sens, Spec, and Acc metrics. However, the Jaccard boxplots show its
high variance compared to the Mini-Unet and the Dense networks. As a consequence, we encourage 
the use of the boxplots to highlights the stability of future models. On all the plots, 
it can be noticed that, for Dense and Mini-Unet, the variations after fine-tuning is 
lower than the ones from patch-level and scratch training.}

\paragraph{When the images are not re-sizable}\rev{ As stated in Sec. \ref{S:ex:db}, the images,
on all databases, are resized before any processing ($e.g.$ patch extraction, pre-processing,
fine-tuning, etc.). However, the metrics are computed after a resize back to the initial size
to have a fair comparison. In our case, the resizing is important for two reasons: $(1)$
to avoid memory overload, and $(2)$ to ease numerical computation when down-sampling  is 
made by the network layers. The resizing did not have a large impact based on our performances
since we are able to reach the state-of-the-art. On tasks where the resizing is not appropriate, our
framework can still be employed by extracting large patches. In other words, suppose the 
initial images are of size $3000\times 3000 \times 3$, one can extract small patches ($e.g.$ 
$64\times 64 \times 3$) to build the patch database, then, extract large patches ($e.g.$ 
$500\times500\times 3$ to build the image database. Therefore, the initial images will
be segmented by aggregating large patches predictions.}
  
\paragraph{A possible limitation} The proposed framework is based on the simple separability assumption, 
that is one can efficiently train a network at the patch-level. 
However, when the extracted patches do not provide enough insight about the 
segmentation task ($e.g.$ they  do not contain any discriminative information) then the previous
assumption may not hold and the transfer to the image-level may fail.

\begin{table}[t]
\caption{Segmentation time on DRIVE (in seconds).}
\label{tab: time}
\begin{center}
 
\centering 
\setlength{\tabcolsep}{0.5em}   
\begin{tabular}{l|l|c|c|}  
\cline{1-4}
&\bf Method &  \bf GPU  & \bf CPU \\
\hline   
\hline 
\parbox[t]{2mm}{\multirow{3}{*}{\rotatebox[origin=c]{90}{P-b}}} &Light & 8.71 & 53  \\
& Mini-Unet & 11.34 & 186 \\
& Dense & 71.47 & 1200 \\  
\hline
\hline
\hline
\parbox[t]{2mm}{\multirow{3}{*}{\rotatebox[origin=c]{90}{I-b}}}&Light &  $\mathbf{.07}$ & $\mathbf{ .25}$  \\
& Mini-Unet & .10  & .84 \\
& Dense & .24  & 4.69 \\  
\hline
\end{tabular}  
\end{center}
\end{table}

\section{Summary and perspectives}
\label{S:sum}
To summarize, this paper presented a training framework for RIS. The proposed framework 
tried to answer  the question \emph{How to obtain state-of-the-art performances at the image-level using a freely designed network on the task of RIS? } To do so, the framework proposes to first freely design a 
network
and train it on patches extracted from the given dataset's images. Then, the knowledge of the previous network is used to 
pre-train the final network at the image-level. The framework was tested on four publicly available datasets  using four different networks.

Another technique when working on small sized databases consists in generating some synthetic samples
using generative adversarial networks (GANs). However, the synthetic data is based on the training 
samples whose size is again relatively small. Future work will include an extensive comparison between
our framework and GAN-based ones. Also, the application of the framework to other image modalities
and its theoretical analysis are part of the next steps we intend to carry.

%% References
%%   \bib{key}          ==>>  [#]

\newpage
\section{References}
 \bibliographystyle{model1-num-names} 
\bibliography{sample.bib}

\begin{thebibliography}{49}
\expandafter\ifx\csname natexlab\endcsname\relax\def\natexlab#1{#1}\fi
\providecommand{\bibinfo}[2]{#2}
\ifx\xfnm\relax \def\xfnm[#1]{\unskip,\space#1}\fi
%Type = Book
\bibitem[{Bowling(2016)}]{Bowling_2016}
\bibinfo{author}{B.~Bowling}, \bibinfo{title}{Kanski's Clinical Ophthalmology:
  A Systematic Approach}, \bibinfo{publisher}{lsevier Health Sciences},
  \bibinfo{edition}{8} edition, \bibinfo{year}{2016}.
%Type = Article
\bibitem[{Abr{\`a}moff~Michael et~al.(2010)Abr{\`a}moff~Michael, Garvin~Mona,
  and Sonka}]{Abramoff_2010}
\bibinfo{author}{D.~Abr{\`a}moff~Michael}, \bibinfo{author}{K.~Garvin~Mona},
  \bibinfo{author}{M.~Sonka},
\newblock \bibinfo{title}{{Retinal imaging and image analysis}},
\newblock \bibinfo{journal}{IEEE reviews in biomedical engineering}
  \bibinfo{volume}{3} (\bibinfo{year}{2010}) \bibinfo{pages}{169--208}.
%Type = Article
\bibitem[{Yau et~al.(2012)Yau, Rogers, Kawasaki, Lamoureux, Kowalski, Bek,
  Chen, Dekker, Fletcher, Grauslund, Haffner, Hamman, Ikram, Kayama, Klein,
  Klein, Krishnaiah, Mayurasakorn, O{\textquoteright}Hare, Orchard, Porta,
  Rema, Roy, Sharma, Shaw, Taylor, Tielsch, Varma, Wang, Wang, West, Xu,
  Yasuda, Zhang, Mitchell, and Wong}]{Yau_2012_Global}
\bibinfo{author}{J.~W. Yau}, \bibinfo{author}{S.~L. Rogers},
  \bibinfo{author}{R.~Kawasaki}, \bibinfo{author}{E.~L. Lamoureux},
  \bibinfo{author}{J.~W. Kowalski}, \bibinfo{author}{T.~Bek},
  \bibinfo{author}{S.-J. Chen}, \bibinfo{author}{J.~M. Dekker},
  \bibinfo{author}{A.~Fletcher}, \bibinfo{author}{J.~Grauslund},
  \bibinfo{author}{S.~Haffner}, \bibinfo{author}{R.~F. Hamman},
  \bibinfo{author}{M.~K. Ikram}, \bibinfo{author}{T.~Kayama},
  \bibinfo{author}{B.~E. Klein}, \bibinfo{author}{R.~Klein},
  \bibinfo{author}{S.~Krishnaiah}, \bibinfo{author}{K.~Mayurasakorn},
  \bibinfo{author}{J.~P. O{\textquoteright}Hare}, \bibinfo{author}{T.~J.
  Orchard}, \bibinfo{author}{M.~Porta}, \bibinfo{author}{M.~Rema},
  \bibinfo{author}{M.~S. Roy}, \bibinfo{author}{T.~Sharma},
  \bibinfo{author}{J.~Shaw}, \bibinfo{author}{H.~Taylor},
  \bibinfo{author}{J.~M. Tielsch}, \bibinfo{author}{R.~Varma},
  \bibinfo{author}{J.~J. Wang}, \bibinfo{author}{N.~Wang},
  \bibinfo{author}{S.~West}, \bibinfo{author}{L.~Xu},
  \bibinfo{author}{M.~Yasuda}, \bibinfo{author}{X.~Zhang},
  \bibinfo{author}{P.~Mitchell}, \bibinfo{author}{T.~Y.~a. Wong},
\newblock \bibinfo{title}{{Global Prevalence and Major Risk Factors of Diabetic
  Retinopathy}},
\newblock \bibinfo{journal}{Diabetes Care} \bibinfo{volume}{35}
  (\bibinfo{year}{2012}) \bibinfo{pages}{556--564}.
%Type = Article
\bibitem[{Tham et~al.(2014)Tham, Li, Wong, Quigley, Aung, and
  Cheng}]{Tham_2014_Global}
\bibinfo{author}{Y.-C. Tham}, \bibinfo{author}{X.~Li}, \bibinfo{author}{T.~Y.
  Wong}, \bibinfo{author}{H.~A. Quigley}, \bibinfo{author}{T.~Aung},
  \bibinfo{author}{C.-Y. Cheng},
\newblock \bibinfo{title}{{Global Prevalence of Glaucoma and Projections of
  Glaucoma Burden through 2040: A Systematic Review and Meta-Analysis}},
\newblock \bibinfo{journal}{Ophthalmology} \bibinfo{volume}{121}
  (\bibinfo{year}{2014}) \bibinfo{pages}{2081--2090}.
%Type = Article
\bibitem[{Fraz et~al.(2012)Fraz, Remagnino, Hoppe, Uyyanonvara, Rudnicka, Owen,
  and Barman}]{Fraz_2012_blood}
\bibinfo{author}{M.~Fraz}, \bibinfo{author}{P.~Remagnino},
  \bibinfo{author}{A.~Hoppe}, \bibinfo{author}{B.~Uyyanonvara},
  \bibinfo{author}{A.~Rudnicka}, \bibinfo{author}{C.~Owen},
  \bibinfo{author}{S.~Barman},
\newblock \bibinfo{title}{{Blood vessel segmentation methodologies in retinal
  images -- A survey}},
\newblock \bibinfo{journal}{Computer Methods and Programs in Biomedicine}
  \bibinfo{volume}{108} (\bibinfo{year}{2012}) \bibinfo{pages}{407--433}.
%Type = Article
\bibitem[{{L Srinidhi} et~al.(2017){L Srinidhi}, Aparna, and
  Rajan}]{Srinidhi_2017_Recent}
\bibinfo{author}{C.~{L Srinidhi}}, \bibinfo{author}{P.~Aparna},
  \bibinfo{author}{J.~Rajan},
\newblock \bibinfo{title}{{Recent Advancements in Retinal Vessel
  Segmentation}},
\newblock \bibinfo{journal}{Journal of Medical Systems} \bibinfo{volume}{41}
  (\bibinfo{year}{2017}) \bibinfo{pages}{70}.
%Type = Article
\bibitem[{Lecun et~al.(1998)Lecun, Bottou, Bengio, and
  Haffner}]{LeCun_1998_Gradient}
\bibinfo{author}{Y.~Lecun}, \bibinfo{author}{L.~Bottou},
  \bibinfo{author}{Y.~Bengio}, \bibinfo{author}{P.~Haffner},
\newblock \bibinfo{title}{Gradient-based learning applied to document
  recognition},
\newblock \bibinfo{journal}{Proceedings of the IEEE} \bibinfo{volume}{86}
  (\bibinfo{year}{1998}) \bibinfo{pages}{2278--2324}.
%Type = Incollection
\bibitem[{Long et~al.(2015)Long, Shelhamer, and Darrell}]{Long_2015_CVPR}
\bibinfo{author}{J.~Long}, \bibinfo{author}{E.~Shelhamer},
  \bibinfo{author}{T.~Darrell},
\newblock \bibinfo{title}{Fully convolutional networks for semantic
  segmentation},
\newblock in: \bibinfo{booktitle}{The IEEE Conference on Computer Vision and
  Pattern Recognition (CVPR)}, \bibinfo{year}{2015}.
%Type = Article
\bibitem[{Litjens et~al.(2017)Litjens, Kooi, Bejnordi, Setio, Ciompi,
  Ghafoorian, van~der Laak, van Ginneken, and
  S{\'a}nchez}]{Litjens_2017_Survey}
\bibinfo{author}{G.~Litjens}, \bibinfo{author}{T.~Kooi}, \bibinfo{author}{B.~E.
  Bejnordi}, \bibinfo{author}{A.~A.~A. Setio}, \bibinfo{author}{F.~Ciompi},
  \bibinfo{author}{M.~Ghafoorian}, \bibinfo{author}{J.~A. van~der Laak},
  \bibinfo{author}{B.~van Ginneken}, \bibinfo{author}{C.~I. S{\'a}nchez},
\newblock \bibinfo{title}{{A survey on deep learning in medical image
  analysis}},
\newblock \bibinfo{journal}{Medical Image Analysis} \bibinfo{volume}{42}
  (\bibinfo{year}{2017}) \bibinfo{pages}{60--88}.
%Type = Book
\bibitem[{Zhou et~al.(2016)Zhou, Greenspan, and Shen}]{Zhou_2017_Deep}
\bibinfo{author}{S.~K. Zhou}, \bibinfo{author}{H.~Greenspan},
  \bibinfo{author}{D.~Shen}, \bibinfo{title}{Deep Learning for Medical Image
  Analysis}, \bibinfo{publisher}{Academic Press}, \bibinfo{year}{(2016)}.
%Type = Article
\bibitem[{Liskowski and Krawiec(2016)}]{Liskowski_2016_Segmenting}
\bibinfo{author}{P.~Liskowski}, \bibinfo{author}{K.~Krawiec},
\newblock \bibinfo{title}{Segmenting retinal blood vessels with deep neural
  networks},
\newblock \bibinfo{journal}{IEEE Transactions on Medical Imaging}
  \bibinfo{volume}{35} (\bibinfo{year}{2016}) \bibinfo{pages}{2369--2380}.
%Type = Article
\bibitem[{Li et~al.(2016)Li, Feng, Xie, Liang, Zhang, and Wang}]{Li_2016_Cross}
\bibinfo{author}{Q.~Li}, \bibinfo{author}{B.~Feng}, \bibinfo{author}{L.~Xie},
  \bibinfo{author}{P.~Liang}, \bibinfo{author}{H.~Zhang},
  \bibinfo{author}{T.~Wang},
\newblock \bibinfo{title}{A cross-modality learning approach for vessel
  segmentation in retinal images},
\newblock \bibinfo{journal}{IEEE Transactions on Medical Imaging}
  \bibinfo{volume}{35} (\bibinfo{year}{2016}) \bibinfo{pages}{109--118}.
%Type = Article
\bibitem[{Tajbakhsh et~al.(2016)Tajbakhsh, Shin, Gurudu, Hurst, Kendall,
  Gotway, and Liang}]{Tajbakhsh_2016_Convolutional}
\bibinfo{author}{N.~Tajbakhsh}, \bibinfo{author}{J.~Y. Shin},
  \bibinfo{author}{S.~R. Gurudu}, \bibinfo{author}{R.~T. Hurst},
  \bibinfo{author}{C.~B. Kendall}, \bibinfo{author}{M.~B. Gotway},
  \bibinfo{author}{J.~Liang},
\newblock \bibinfo{title}{Convolutional neural networks for medical image
  analysis: Full training or fine tuning?},
\newblock \bibinfo{journal}{IEEE Transactions on Medical Imaging}
  \bibinfo{volume}{35} (\bibinfo{year}{2016}) \bibinfo{pages}{1299--1312}.
%Type = Article
\bibitem[{Yosinski et~al.(2014)Yosinski, Clune, Bengio, and
  Lipson}]{Yosinski_2014_How}
\bibinfo{author}{J.~Yosinski}, \bibinfo{author}{J.~Clune},
  \bibinfo{author}{Y.~Bengio}, \bibinfo{author}{H.~Lipson},
\newblock \bibinfo{title}{{How transferable are features in deep neural
  networks?}}  (\bibinfo{year}{2014}) \bibinfo{pages}{3320--3328}.
%Type = Article
\bibitem[{Szegedy et~al.(2014)Szegedy, Liu, Jia, Sermanet, Reed, Anguelov,
  Erhan, Vanhoucke, and Rabinovich}]{Szegedy_2014_Going}
\bibinfo{author}{C.~Szegedy}, \bibinfo{author}{W.~Liu},
  \bibinfo{author}{Y.~Jia}, \bibinfo{author}{P.~Sermanet},
  \bibinfo{author}{S.~E. Reed}, \bibinfo{author}{D.~Anguelov},
  \bibinfo{author}{D.~Erhan}, \bibinfo{author}{V.~Vanhoucke},
  \bibinfo{author}{A.~Rabinovich},
\newblock \bibinfo{title}{Going deeper with convolutions},
\newblock \bibinfo{journal}{CoRR} \bibinfo{volume}{abs/1409.4842}
  (\bibinfo{year}{2014}).
%Type = Article
\bibitem[{Krizhevsky et~al.(2012)Krizhevsky, Sutskever, and
  Hinton}]{Krizhevsky_2012_ImageNet}
\bibinfo{author}{A.~Krizhevsky}, \bibinfo{author}{I.~Sutskever},
  \bibinfo{author}{G.~E. Hinton},
\newblock \bibinfo{title}{Imagenet classification with deep convolutional
  neural networks}  (\bibinfo{year}{2012}) \bibinfo{pages}{1097--1105}.
%Type = Article
\bibitem[{He et~al.(2015)He, Zhang, Ren, and Sun}]{He_2015_Deep}
\bibinfo{author}{K.~He}, \bibinfo{author}{X.~Zhang}, \bibinfo{author}{S.~Ren},
  \bibinfo{author}{J.~Sun},
\newblock \bibinfo{title}{Deep residual learning for image recognition},
\newblock \bibinfo{journal}{CoRR} \bibinfo{volume}{abs/1512.03385}
  (\bibinfo{year}{2015}).
%Type = Article
\bibitem[{Huang et~al.(2017)Huang, Liu, van~der Maaten, and
  Weinberger}]{Huang_2017_Densely}
\bibinfo{author}{G.~Huang}, \bibinfo{author}{Z.~Liu},
  \bibinfo{author}{L.~van~der Maaten}, \bibinfo{author}{K.~Q. Weinberger},
\newblock \bibinfo{title}{Densely connected convolutional networks},
\newblock \bibinfo{journal}{The IEEE Conference on Computer Vision and Pattern
  Recognition (CVPR)}  (\bibinfo{year}{2017}).
%Type = Article
\bibitem[{Saxena and Verbeek(2016)}]{Saxena_2016_Convolutional}
\bibinfo{author}{S.~Saxena}, \bibinfo{author}{J.~Verbeek},
\newblock \bibinfo{title}{{Convolutional Neural Fabrics}},
\newblock \bibinfo{journal}{{Advances in Neural Information Processing Systems
  29}}  (\bibinfo{year}{2016}) \bibinfo{pages}{4053--4061}.
%Type = Incollection
\bibitem[{V{\'e}niat and Denoyer(2018)}]{Veniat_2018_CVPR}
\bibinfo{author}{T.~V{\'e}niat}, \bibinfo{author}{L.~Denoyer},
\newblock \bibinfo{title}{Learning time/memory-efficient deep architectures
  with budgeted super networks},
\newblock in: \bibinfo{booktitle}{The IEEE Conference on Computer Vision and
  Pattern Recognition (CVPR)}, \bibinfo{year}{2018}.
%Type = Article
\bibitem[{Ronneberger et~al.(2015)Ronneberger, Fischer, and Brox}]{UNET}
\bibinfo{author}{O.~Ronneberger}, \bibinfo{author}{P.~Fischer},
  \bibinfo{author}{T.~Brox},
\newblock \bibinfo{title}{{U-Net: Convolutional Networks for Biomedical Image
  Segmentation}},
\newblock \bibinfo{journal}{{Medical Image Computing and Computer-Assisted
  Intervention -- MICCAI 2015}}  (\bibinfo{year}{2015})
  \bibinfo{pages}{234--241}.
%Type = Article
\bibitem[{Moccia et~al.(2018)Moccia, Momi, Hadji, and
  Mattos}]{Moccia_2018_Blood}
\bibinfo{author}{S.~Moccia}, \bibinfo{author}{E.~D. Momi},
  \bibinfo{author}{S.~E. Hadji}, \bibinfo{author}{L.~S. Mattos},
\newblock \bibinfo{title}{{Blood vessel segmentation algorithms --- Review of
  methods, datasets and evaluation metrics}},
\newblock \bibinfo{journal}{Computer Methods and Programs in Biomedicine}
  \bibinfo{volume}{158} (\bibinfo{year}{2018}) \bibinfo{pages}{71--91}.
%Type = Article
\bibitem[{Hajabdollahi et~al.(2018)Hajabdollahi, Esfandiarpoor, Najarian,
  Karimi, Samavi, and Reza-Soroushmeh}]{Hajabdollahi_2018}
\bibinfo{author}{M.~Hajabdollahi}, \bibinfo{author}{R.~Esfandiarpoor},
  \bibinfo{author}{K.~Najarian}, \bibinfo{author}{N.~Karimi},
  \bibinfo{author}{S.~Samavi}, \bibinfo{author}{S.~M. Reza-Soroushmeh},
\newblock \bibinfo{title}{Low complexity convolutional neural network for
  vessel segmentation in portable retinal diagnostic devices},
\newblock \bibinfo{journal}{25th IEEE International Conference on Image
  Processing (ICIP)}  (\bibinfo{year}{2018}) \bibinfo{pages}{2785--2789}.
%Type = Article
\bibitem[{Oliveira et~al.(2018)Oliveira, Pereira, and
  Silva}]{Oliveira_1018_Fully}
\bibinfo{author}{A.~Oliveira}, \bibinfo{author}{S.~Pereira},
  \bibinfo{author}{C.~A. Silva},
\newblock \bibinfo{title}{Retinal vessel segmentation based on fully
  convolutional neural networks},
\newblock \bibinfo{journal}{Expert Systems with Applications}
  \bibinfo{volume}{112} (\bibinfo{year}{2018}) \bibinfo{pages}{229 -- 242}.
%Type = Article
\bibitem[{Jiang et~al.(2018)Jiang, Zhang, Wang, and Ko}]{Jiang_2018_Retinal}
\bibinfo{author}{Z.~Jiang}, \bibinfo{author}{H.~Zhang},
  \bibinfo{author}{Y.~Wang}, \bibinfo{author}{S.-B. Ko},
\newblock \bibinfo{title}{Retinal blood vessel segmentation using fully
  convolutional network with transfer learning},
\newblock \bibinfo{journal}{Computerized Medical Imaging and Graphics}
  \bibinfo{volume}{68} (\bibinfo{year}{2018}) \bibinfo{pages}{1 -- 15}.
%Type = Article
\bibitem[{Dasgupta and Singh(2017)}]{Dasgupta_2017_Fully}
\bibinfo{author}{A.~Dasgupta}, \bibinfo{author}{S.~Singh},
\newblock \bibinfo{title}{A fully convolutional neural network based structured
  prediction approach towards the retinal vessel segmentation},
\newblock \bibinfo{journal}{The IEEE 14th International Symposium on Biomedical
  Imaging (ISBI)}  (\bibinfo{year}{2017}) \bibinfo{pages}{248--251}.
%Type = Article
\bibitem[{Yao(2017)}]{Yang_2017_Patch}
\bibinfo{author}{J.~Y.~L. Yao},
\newblock \bibinfo{title}{Patch-based fully convolutional neural network with
  skip connections for retinal blood vessel segmentation},
\newblock \bibinfo{journal}{24th IEEE International Conference on Image
  Processing (ICIP)}  (\bibinfo{year}{2017}).
%Type = Article
\bibitem[{Fu et~al.(2018)Fu, Cheng, Xu, {Kee Wong}, Liu, and
  Cao}]{Fu_2018_Joint}
\bibinfo{author}{H.~Fu}, \bibinfo{author}{J.~Cheng}, \bibinfo{author}{Y.~Xu},
  \bibinfo{author}{D.~W. {Kee Wong}}, \bibinfo{author}{J.~Liu},
  \bibinfo{author}{X.~Cao},
\newblock \bibinfo{title}{Joint optic disc and cup segmentation based on
  multi-label deep network and polar transformation},
\newblock \bibinfo{journal}{IEEE Transactions on Medical Imaging (TMI)}
  \bibinfo{volume}{37} (\bibinfo{year}{2018}) \bibinfo{pages}{1597--1605}.
%Type = Article
\bibitem[{Tan et~al.(2017)Tan, Acharya, Bhandary, Chua, and
  Sivaprasad}]{Tan_2017_Optic}
\bibinfo{author}{J.~H. Tan}, \bibinfo{author}{U.~R. Acharya},
  \bibinfo{author}{S.~V. Bhandary}, \bibinfo{author}{K.~C. Chua},
  \bibinfo{author}{S.~Sivaprasad},
\newblock \bibinfo{title}{{Segmentation of optic disc, fovea and retinal
  vasculature using a single convolutional neural network}},
\newblock \bibinfo{journal}{Journal of Computational Science}
  \bibinfo{volume}{20} (\bibinfo{year}{2017}) \bibinfo{pages}{70--79}.
%Type = Article
\bibitem[{Fu et~al.(2016)Fu, Xu, Wong, and Liu}]{Fu_2016}
\bibinfo{author}{H.~Fu}, \bibinfo{author}{Y.~Xu}, \bibinfo{author}{D.~W.~K.
  Wong}, \bibinfo{author}{J.~Liu},
\newblock \bibinfo{title}{Retinal vessel segmentation via deep learning network
  and fully-connected conditional random fields},
\newblock \bibinfo{journal}{The IEEE 13th International Symposium on Biomedical
  Imaging (ISBI)}  (\bibinfo{year}{2016}) \bibinfo{pages}{698--701}.
%Type = Article
\bibitem[{Xie and Tu(2017)}]{HED}
\bibinfo{author}{S.~Xie}, \bibinfo{author}{Z.~Tu},
\newblock \bibinfo{title}{{Holistically-Nested Edge Detection}},
\newblock \bibinfo{journal}{International Journal of Computer Vision}
  \bibinfo{volume}{125} (\bibinfo{year}{2017}) \bibinfo{pages}{3--18}.
%Type = Article
\bibitem[{Simonyan and Zisserman(2014)}]{Simonyan_2014_Very}
\bibinfo{author}{K.~Simonyan}, \bibinfo{author}{A.~Zisserman},
\newblock \bibinfo{title}{Very deep convolutional networks for large-scale
  image recognition},
\newblock \bibinfo{journal}{CoRR} \bibinfo{volume}{abs/1409.1556}
  (\bibinfo{year}{2014}).
%Type = Article
\bibitem[{Mo and Zhang(2017)}]{Mo_2017_Multi}
\bibinfo{author}{J.~Mo}, \bibinfo{author}{L.~Zhang},
\newblock \bibinfo{title}{{Multi-level deep supervised networks for retinal
  vessel segmentation}},
\newblock \bibinfo{journal}{International Journal of Computer Assisted
  Radiology and Surgery} \bibinfo{volume}{12} (\bibinfo{year}{2017})
  \bibinfo{pages}{2181--2193}.
%Type = Article
\bibitem[{Maninis et~al.(2016)Maninis, Pont-Tuset, Arbel{\'a}ez, and {Van
  Gool}}]{Maninis_2016}
\bibinfo{author}{K.-K. Maninis}, \bibinfo{author}{J.~Pont-Tuset},
  \bibinfo{author}{P.~Arbel{\'a}ez}, \bibinfo{author}{L.~{Van Gool}},
\newblock \bibinfo{title}{{Deep Retinal Image Understanding}},
\newblock \bibinfo{journal}{Medical Image Computing and Computer-Assisted
  Intervention (MICCAI)}  (\bibinfo{year}{2016}) \bibinfo{pages}{140--148}.
%Type = Article
\bibitem[{{Birgui Sekou} et~al.(2018){Birgui Sekou}, Hidane, Olivier, and
  Cardot}]{Birgui_2018_Retinal}
\bibinfo{author}{T.~{Birgui Sekou}}, \bibinfo{author}{M.~Hidane},
  \bibinfo{author}{J.~Olivier}, \bibinfo{author}{H.~Cardot},
\newblock \bibinfo{title}{{Retinal Blood Vessel Segmentation Using a Fully
  Convolutional Network -- Transfer Learning from Patch- to Image-Level}},
\newblock \bibinfo{journal}{In Machine Learning in Medical Imaging (MLMI)}
  (\bibinfo{year}{2018}) \bibinfo{pages}{170--178}.
%Type = Article
\bibitem[{LeCun et~al.(1989)LeCun, Boser, Denker, Henderson, Howard, Hubbard,
  and Jackel}]{LeCun_1989_Backpropagation}
\bibinfo{author}{Y.~LeCun}, \bibinfo{author}{B.~Boser}, \bibinfo{author}{J.~S.
  Denker}, \bibinfo{author}{D.~Henderson}, \bibinfo{author}{R.~E. Howard},
  \bibinfo{author}{W.~Hubbard}, \bibinfo{author}{L.~D. Jackel},
\newblock \bibinfo{title}{{Backpropagation Applied to Handwritten Zip Code
  Recognition}},
\newblock \bibinfo{journal}{Neural Computation} \bibinfo{volume}{1}
  (\bibinfo{year}{1989}) \bibinfo{pages}{541--551}.
%Type = Article
\bibitem[{Goodfellow et~al.(2014)Goodfellow, Pouget-Abadie, Mirza, Xu,
  Warde-Farley, Ozair, Courville, and Bengio}]{Goodfellow_2014_Generative}
\bibinfo{author}{I.~Goodfellow}, \bibinfo{author}{J.~Pouget-Abadie},
  \bibinfo{author}{M.~Mirza}, \bibinfo{author}{B.~Xu},
  \bibinfo{author}{D.~Warde-Farley}, \bibinfo{author}{S.~Ozair},
  \bibinfo{author}{A.~Courville}, \bibinfo{author}{Y.~Bengio},
\newblock \bibinfo{title}{Generative adversarial nets},
\newblock \bibinfo{journal}{{Advances in Neural Information Processing Systems
  27}}  (\bibinfo{year}{2014}) \bibinfo{pages}{2672--2680}.
%Type = Book
\bibitem[{Goodfellow et~al.(2016)Goodfellow, Bengio, and
  Courville}]{Goodfellow_2016_Deep}
\bibinfo{author}{I.~Goodfellow}, \bibinfo{author}{Y.~Bengio},
  \bibinfo{author}{A.~Courville}, \bibinfo{title}{Deep Learning},
  \bibinfo{publisher}{MIT Press}, \bibinfo{year}{(2016)}.
%Type = Article
\bibitem[{LeCun and Hinton(2015)}]{LeCun_2015_Deep}
\bibinfo{author}{B.~Y. LeCun, Y.}, \bibinfo{author}{G.~E. Hinton},
\newblock \bibinfo{title}{{Deep Learning}},
\newblock \bibinfo{journal}{Nature} \bibinfo{volume}{521}
  (\bibinfo{year}{2015}) \bibinfo{pages}{436--444}.
%Type = Article
\bibitem[{Zeiler and Fergus(2014)}]{Zeiler_2014_Visualizing}
\bibinfo{author}{M.~D. Zeiler}, \bibinfo{author}{R.~Fergus},
\newblock \bibinfo{title}{{Visualizing and Understanding Convolutional
  Networks}},
\newblock \bibinfo{journal}{{In European Conference on Computer Vision (ECCV)}}
   (\bibinfo{year}{2014}) \bibinfo{pages}{818--833}.
%Type = Article
\bibitem[{Pan and Yang(2010)}]{Pan_2010_Survey}
\bibinfo{author}{S.~J. Pan}, \bibinfo{author}{Q.~Yang},
\newblock \bibinfo{title}{A survey on transfer learning},
\newblock \bibinfo{journal}{IEEE Transactions on Knowledge and Data
  Engineering} \bibinfo{volume}{22} (\bibinfo{year}{2010})
  \bibinfo{pages}{1345--1359}.
%Type = Article
\bibitem[{Zhou et~al.(2017)Zhou, Shin, Zhang, Gurudu, Gotway, and
  Liang}]{Zhou_2017_CVPR}
\bibinfo{author}{Z.~Zhou}, \bibinfo{author}{J.~Shin},
  \bibinfo{author}{L.~Zhang}, \bibinfo{author}{S.~Gurudu},
  \bibinfo{author}{M.~Gotway}, \bibinfo{author}{J.~Liang},
\newblock \bibinfo{title}{Fine-tuning convolutional neural networks for
  biomedical image analysis: Actively and incrementally},
\newblock \bibinfo{journal}{The IEEE Conference on Computer Vision and Pattern
  Recognition (CVPR)}  (\bibinfo{year}{2017}).
%Type = Article
\bibitem[{{Zoran} and {Weiss}(2011)}]{Zoran_2011}
\bibinfo{author}{D.~{Zoran}}, \bibinfo{author}{Y.~{Weiss}},
\newblock \bibinfo{title}{From learning models of natural image patches to
  whole image restoration},
\newblock \bibinfo{journal}{International Conference on Computer Vision}
  (\bibinfo{year}{2011}) \bibinfo{pages}{479--486}.
%Type = Article
\bibitem[{Staal et~al.(2004)Staal, Abramoff, Niemeijer, Viergever, and van
  Ginneken}]{DRIVE}
\bibinfo{author}{J.~Staal}, \bibinfo{author}{M.~Abramoff},
  \bibinfo{author}{M.~Niemeijer}, \bibinfo{author}{M.~Viergever},
  \bibinfo{author}{B.~van Ginneken},
\newblock \bibinfo{title}{{Ridge based vessel segmentation in color images of
  the retina}},
\newblock \bibinfo{journal}{{IEEE Transactions on Medical Imaging}}
  \bibinfo{volume}{23} (\bibinfo{year}{2004}) \bibinfo{pages}{501--509}.
%Type = Article
\bibitem[{Hoover and Goldbaum(2003)}]{STARE}
\bibinfo{author}{A.~W. Hoover}, \bibinfo{author}{M.~H. Goldbaum},
\newblock \bibinfo{title}{{Locating the Optical Nerve in a Retinal Image Using
  the Fuzzy Convergence of the Blood Vessels.}},
\newblock \bibinfo{journal}{IEEE Trans. Med. Imaging} \bibinfo{volume}{22}
  (\bibinfo{year}{2003}) \bibinfo{pages}{951--958}.
%Type = Article
\bibitem[{Owen et~al.(2009)Owen, Rudnicka, Mullen, Barman, Monekosso, Whincup,
  Ng, and Paterson}]{CHASE}
\bibinfo{author}{C.~G. Owen}, \bibinfo{author}{A.~R. Rudnicka},
  \bibinfo{author}{R.~Mullen}, \bibinfo{author}{S.~A. Barman},
  \bibinfo{author}{D.~Monekosso}, \bibinfo{author}{P.~H. Whincup},
  \bibinfo{author}{J.~Ng}, \bibinfo{author}{C.~Paterson},
\newblock \bibinfo{title}{{Measuring Retinal Vessel Tortuosity in 10-Year-Old
  Children: Validation of the Computer-Assisted Image Analysis of the Retina
  (CAIAR) Program}},
\newblock \bibinfo{journal}{Investigative Ophthalmology and Visual Science}
  \bibinfo{volume}{50} (\bibinfo{year}{2009}) \bibinfo{pages}{2004--2010}.
%Type = Article
\bibitem[{Porwal et~al.(2018)Porwal, Pachade, Kamble, Kokare, Deshmukh,
  Sahasrabuddhe, and Meriaudeau}]{IDRID}
\bibinfo{author}{P.~Porwal}, \bibinfo{author}{S.~Pachade},
  \bibinfo{author}{R.~Kamble}, \bibinfo{author}{M.~Kokare},
  \bibinfo{author}{G.~Deshmukh}, \bibinfo{author}{V.~Sahasrabuddhe},
  \bibinfo{author}{F.~Meriaudeau},
\newblock \bibinfo{title}{Indian diabetic retinopathy image dataset (idrid)},
\newblock \bibinfo{journal}{IEEE Dataport}  (\bibinfo{year}{2018}).
%Type = Article
\bibitem[{Zeiler(2012)}]{Zeiler_2012_ADADELTA}
\bibinfo{author}{M.~D. Zeiler},
\newblock \bibinfo{title}{{ADADELTA:} an adaptive learning rate method},
\newblock \bibinfo{journal}{CoRR} \bibinfo{volume}{abs/1212.5701}
  (\bibinfo{year}{2012}).
%Type = Article
\bibitem[{He et~al.(2017)He, Gkioxari, Doll{\'{a}}r, and
  Girshick}]{He_MaskRCNN}
\bibinfo{author}{K.~He}, \bibinfo{author}{G.~Gkioxari},
  \bibinfo{author}{P.~Doll{\'{a}}r}, \bibinfo{author}{R.~B. Girshick},
\newblock \bibinfo{title}{Mask {R-CNN}},
\newblock \bibinfo{journal}{CoRR} \bibinfo{volume}{abs/1703.06870}
  (\bibinfo{year}{2017}).

\end{thebibliography}






% \end{thebibliography}

\end{document}